\documentclass{article}

\PassOptionsToPackage{numbers, compress}{natbib}

 \usepackage[preprint]{neurips_2026}

\usepackage[utf8]{inputenc} 
\usepackage[T1]{fontenc}    
\usepackage{hyperref}       
\usepackage{url}            
\usepackage{booktabs}       
\usepackage{amsfonts}       
\usepackage{nicefrac}       
\usepackage{microtype}      
\usepackage{xcolor}         
\usepackage{hyperref}
\usepackage{multicol}
\usepackage{multirow}
\usepackage{enumitem}
\usepackage{subcaption}
\usepackage{makecell}
\usepackage{graphicx}
\usepackage{amsmath}
\newtheorem{definition}{Definition}
\usepackage{algorithm}
\usepackage{algpseudocode}
\usepackage{xspace}

\newcommand{\pgraph}[1]{\textbf{#1}\,\,}

\newcommand{\rev}[1]{{#1}}

\def\ftfm{\texttt{\texttt{FairTFM}}\xspace}
\title{Training Fair Tabular Foundation Models}

\author{%
  Patrik Kenfack \\
  \'ETS Montr\'eal \\
  Mila - Quebec AI Institute\\
  \small\texttt{patrik-joslin.kenfack.1@ens.etsmtl.ca} \\
  \And
    Jesse C. Cresswell\\
  Layer 6 AI \\
  Toronto, Canada\\
  \small\texttt{jesse@layer6.ai} \\
    \And
    Anthony L. Caterini\\
  Layer 6 AI \\
  Toronto, Canada\\
  \small\texttt{anthony@layer6.ai} \\
  \And  
  Samira Ebrahimi Kahou\\
  University of Calgary \\
  Mila - Quebec AI Institute, CIFAR\\
  \small\texttt{samira.ebrahimikahou@ucalgary.ca} \\
    \And  
  Ulrich A\"ivodji \\
  \'ETS Montr\'eal \\
  Mila - Quebec AI Institute\\
  \small\texttt{ulrich.aivodji@etsmtl.ca} \\
}

\begin{document}

\maketitle

\begin{abstract}
Tabular Foundation Models (TFMs) have emerged as leading methods for tabular predictive tasks, leveraging in-context learning to predict on new data without task-specific training. Despite the increased use of TFMs in high-stakes decision-making, their fairness properties remain largely unexplored. In this work, we incorporate fairness constraints directly into TFM training, enabling fair predictions in a single forward pass.
Our approach addresses two key challenges: limited access to sensitive attributes in training data, and the incompatibility of existing fairness techniques with the in-context learning paradigm. We propose \ftfm, a scalable training strategy based on synthetic fairness tasks and a fairness-aware architecture using a gradient reversal layer, which encourages the model to learn representations invariant to sensitive attributes. Experiments on 132 fairness tasks show consistent improvements in fairness while maintaining competitive accuracy. 
\end{abstract}

\section{Introduction}

Tabular data, organized in rows and columns, is the dominant modality for decision-making tasks in domains such as healthcare and finance. While tree-based models, such as XGBoost~\citep{chen2016xgboost}, have long dominated tabular learning, tabular foundation models (TFMs) have emerged as strong alternatives~\citep{hollmanntabpfn}. These models are typically pretrained on large collections of tabular data and can adapt to new tasks via in-context learning (ICL)~\citep{brown2020language}, using only a few labeled examples. Unlike traditional approaches, TFMs do not require task-specific training or hyperparameter tuning. Notably, models such as TabDPT~\citep{ma2024tabdpt}, TabPFNv2~\citep{hollmann2025accurate}, and TabICLv2~\citep{qu2026tabiclv2} can match or outperform heavily tuned tree-based models across a wide range of datasets~\citep{erickson2025tabarena}.

Even as TFMs are deployed in high-stakes decision-making, their fairness properties remain largely unexplored. Recent work shows that, despite strong predictive performance, TFMs can exhibit biased outcomes similar to traditional models~\citep{kenfack2026towards}. While fairness-aware training has been extensively studied in classical machine learning~\citep{mehrabi2021survey}, these approaches do not readily extend to the ICL paradigm, where predictions must be produced in a single forward pass without task-specific optimization. This gap calls for fairness-aware training methods specifically designed for TFMs. Hence, in this work, we propose a training framework that incorporates fairness as a first-class objective in TFMs alongside predictive performance. We train TFMs under explicit fairness constraints, enabling fair predictions directly through ICL without requiring post-hoc correction or task-specific retraining. To our knowledge, the resulting model is the first TFM that accounts for statistical group fairness notions, such as demographic parity and equalized odds, during pretraining.

\looseness=-1 As a running example, consider a bank that uses a TFM to predict whether a client's income will exceed \$50,000. Anti-discrimination regulations forbid disadvantaging clients based on gender, which could occur downstream if the income predictions are biased. Since an off-the-shelf TFM makes predictions without modification to its weights, there is no opportunity to apply fairness constraints specific to this task. With our approach, the bank simply marks gender as the sensitive attribute in the context set, and the pretrained model produces fairer predictions in a single forward pass.

Our method for fair pretraining relies on two key components. (i) \textbf{Synthetic fairness task generation:} given a dataset, we randomly designate an input feature as a sensitive attribute, treat it as categorical, and optimize for fairness with respect to it. Repeating this process for every dataset sampled during pretraining enables scalable fairness-aware training across diverse tasks. (ii) \textbf{Fairness-aware architecture and training:} we extend transformer-based TFMs with a dedicated encoder for sensitive attributes and introduce a dual-head prediction mechanism. In addition to the label prediction head, we include a sensitive attribute predictor connected through a gradient reversal layer~\citep{ganin2016domain}, encouraging the model to learn representations invariant to sensitive attributes.

\looseness=-1
We evaluate our approach on 120 fairness tasks derived from the ACS PUMS datasets~\citep{ding2021retiring} with varying sensitive attributes, and on 12 additional tasks built from six widely used fairness benchmarks to test generalization. We measure fairness using demographic parity, equal opportunity, and equalized odds, comparing against strong baselines. Our results show that the proposed \ftfm framework consistently improves fairness metrics while maintaining competitive predictive performance, improving fairness by 32--75\% relative to the strongest TFM baseline at accuracy costs of 2--11\%.

Our contributions are as follows:

\begin{itemize}[leftmargin=*, nosep]
    \item We propose a scalable strategy for creating fairness tasks with a wide variety of sensitive attribute relationships, based on a stick-breaking group construction that requires no labelled sensitive attributes.
    \item We design a fairness-aware transformer architecture using adversarial learning that enables fair predictions in a single forward pass, handling a sensitive attribute that changes identity across tasks and is masked at inference.
    \item We provide an extensive empirical evaluation on a diverse set of tasks, demonstrating improved fairness metrics while preserving accuracy, including comparisons with fairness-aware baselines and probing analyses of the learned representations.
\end{itemize}
 
\section{Related work}
\pgraph{Fairness} Prior work on fair machine learning typically intervenes at one of three stages: pre-processing the data, incorporating fairness constraints during training, or post-processing model outputs~\citep{kamiran2009classifying,zemel2013learning,hardt2016equality, zhang2018mitigating}. These approaches have been studied extensively for conventional supervised models, but they often assume access to the training pipeline or to calibrated model outputs. This assumption is less compatible with ICL, where a pretrained model is used as a frozen predictor at inference time. Our setting is therefore closer to learning fairness-aware representations during pretraining so that fair behaviour can be obtained in a single forward pass. While adversarial fairness objectives based on gradient reversal have previously been studied in unsupervised domain adaptation~\cite{ganin2015grl}, integrating such objectives into TFM pretraining introduces fundamentally different challenges, including heterogeneous tasks, dynamically changing sensitive attributes across sampled tasks, and the requirement of single-pass in-context inference without task-specific optimization. Our contribution is therefore not merely the use of adversarial learning for fairness, but the introduction of a scalable fairness-aware pretraining framework for TFMs that enables fairer predictions across unseen downstream tasks while preserving the inference properties of in-context learning.

\pgraph{Tabular foundation models} Recent tabular foundation models such as TabPFN, TabDPT, and TabICL demonstrate that pretrained transformers can be highly competitive on tabular prediction tasks~\citep{hollmanntabpfn,hollmann2025accurate,ma2024tabdpt, grinsztajn2025tabpfn,qu2025tabicl,qu2026tabiclv2}. However, this literature emphasizes predictive accuracy, and does not directly address the potential for biased outcomes. FairPFN~\citep{robertsonfairpfn} does incorporate fairness into TFM pretraining under a causal notion of fairness, which contrasts with our work which targets statistical group fairness notions such as demographic parity, equal opportunity, and equalized odds. Statistical fairness is data-driven and focuses on distributional fairness (outcome-based), while causal fairness is more interventional and does not necessarily ensure outcome parity across groups. Our work focuses directly on optimizing statistical fairness notions commonly used in tabular classification benchmarks~\cite{le2022survey}. \rev{We include FairPFN as a baseline in our experiments and find that its causal objective does yield partial group-fairness benefits, but at a predictive cost that leaves it Pareto-dominated by \ftfm{}, and without any mechanism for controlling the fairness--utility trade-off.} We provide a more detailed discussion of these connections in Appendix~\ref{app:related-work}.

\section{Training fair tabular foundation models}
\label{sec:method}
In this section, we present our fairness-aware tabular foundation model for statistical fairness. We describe how fairness tasks are synthesized from generic data priors, introduce the model architecture, and specify the pretraining objective.   

\begin{figure}[t]
    \centering
    \begin{subfigure}{0.54\textwidth}
        \centering
        \includegraphics[width=\linewidth, trim={90 20 40 60}, clip]{figures/figure_step_1.pdf}
        \caption{Fairness task sampling from a prior.}
        \label{fig:overview:sub1}
    \end{subfigure}
    \begin{subfigure}{0.45\textwidth}
        \centering
        \includegraphics[width=\linewidth, trim={150 36 200 30}, clip]{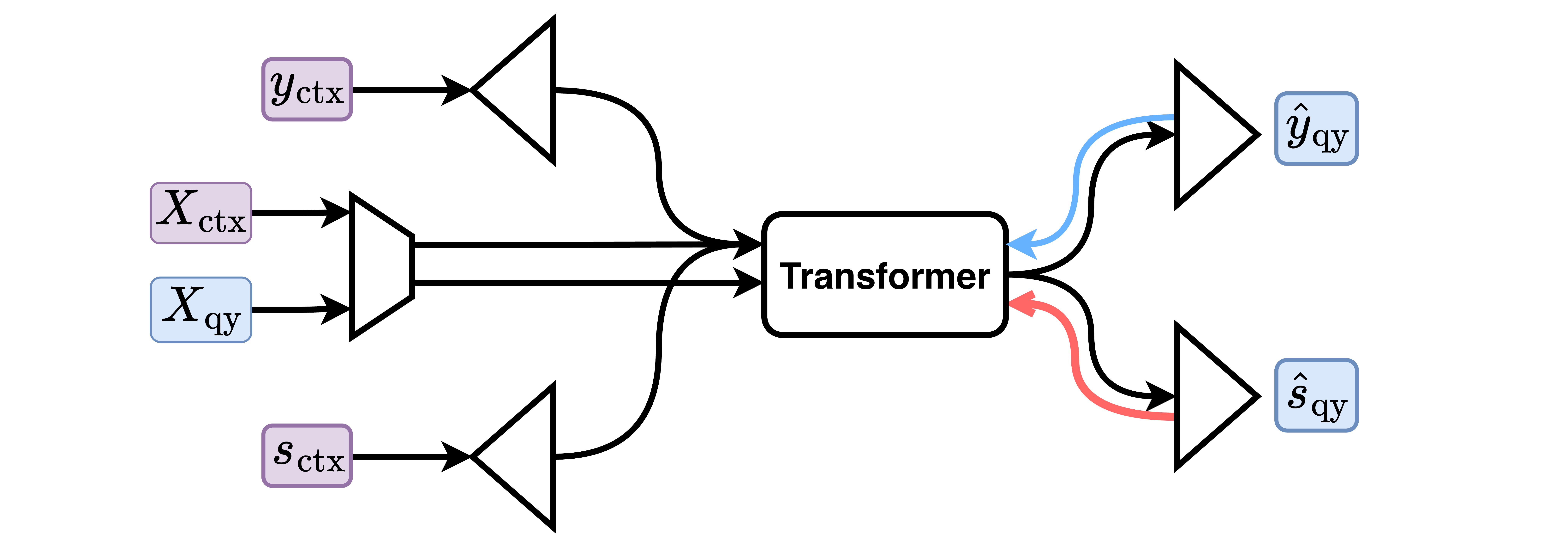}
        \caption{Overview of the \ftfm{} architecture.}
        \label{fig:overview:sub2}
    \end{subfigure}
    
   \caption{(a) We construct fairness tasks from a prior data generator by first sampling a dataset $(X',y)$, then randomly removing one feature to serve as the sensitive attribute $s$ and converting it to a categorical variable, yielding the triplet $(X,y,s)$. The resulting data are split into context $(X_{\text{ctx}},y_{\text{ctx}},s_{\text{ctx}})$ and query $(X_{\text{qy}},y_{\text{qy}},s_{\text{qy}})$ sets for ICL. (b) \ftfm jointly processes context and query inputs with a transformer. In addition to the main prediction head for the target label $\hat{y}_{\text{qy}}$, a second head predicts the sensitive attribute $\hat{s}_{\text{qy}}$. A gradient reversal layer on the sensitive branch (red arrow) encourages the model to learn representations that are invariant to $s$, thereby reducing reliance on sensitive-correlated features while preserving predictive performance.}
    \label{fig:overview}
\end{figure}

\subsection{Fairness task sampling from a dataset prior}\label{sec:fairness_task_sampling} Most TFMs are pretrained on tasks sampled from a dataset prior, whether synthetic or representing real datasets, in order to expose the model to a broad distribution of supervised learning problems. We follow the same principle, but augment it to generate \emph{fairness tasks} in a self-supervised manner~\cite{chen2020simclr, balestriero2023cookbook} that captures diverse forms of group-dependent bias. A similar process was employed in TabDPT~\cite{ma2024tabdpt} for crafting diverse predictive tasks from real-world datasets, and in CausalPFN~\cite{balazadeh2025causalpfn, stith2026causal} for generating synthetic causal inference tasks.

Concretely, starting from a sampled dataset, we randomly designate one input feature as the sensitive attribute \rev{and remove it from the feature set}, as illustrated in Figure~\ref{fig:overview:sub1}. When the selected feature is continuous, we convert it into a categorical attribute using a ``stick-breaking'' discretization scheme. We first sample a sequence of proportions from Beta distributions and use them to construct mixture weights via a Dirichlet process~\citep{khan2012stick}, where each component represents a fraction of the remaining mass. These weights are then mapped to empirical quantiles of the normalized feature, producing data-dependent cut points that partition the feature into discrete groups. This procedure yields flexible, non-uniform bins that adapt to the underlying feature distribution while retaining a probabilistic interpretation.

\rev{In the running example, the bank's dataset corresponds to one predictive task requiring fairness considerations, where gender plays the role of the designated sensitive attribute. During pretraining, any feature in a dataset can play the role of sensitive attribute in some sampled tasks, and an ordinary predictive feature in others.}

\subsection{Network architecture}
We build on TabPFN's transformer encoder for tabular data~\citep{vaswani2017attention}, using the nanoTabPFN architecture~\citep{pfefferle_nanotabpfn_2025} as a lightweight backbone that also has less input pre-processing so that we mitigate confounding effects in our controlled experiments. In this architecture, each input pair $(X_i, y_i)$ is represented as a sequence of $d$-dimensional tokens and is processed by alternating self-attention over rows and columns, enabling ICL along both axes. 

During pretraining, available input data is divided into context $(X_{\mathrm{ctx}},y_{\mathrm{ctx}})$ and query $(X_{\mathrm{qy}},y_{\mathrm{qy}})$ sets. As in TabPFN, the query labels $y_{\mathrm{qy}}$ are masked and the corresponding output token is mapped to class logits through a multi-layer perceptron (MLP). For inference, available labelled data (what would be training data for a non-foundational predictive model) is provided as context, and the TFM predicts $y_{\mathrm{qy}}$ on unlabelled query data.

Our fairness-aware extension, which we call Fair Tabular Foundation Model (\ftfm), is shown in Figure~\ref{fig:overview:sub2}. Rather than operating only on pairs $(x_i,y_i)$, \ftfm processes triplets $(x_i,y_i,s_i)$, where $s_i$ denotes the $d$-dimensional token of the sensitive attribute. \rev{\ftfm incorporates a shared transformer backbone that learns representations, and two predictive MLP heads for the target label $\hat{y}_{\text{qy}}$ and sensitive attribute $\hat{s}_{\text{qy}}$. 
Because the chosen attention mechanism uses no column positional encodings, predictions are invariant to the ordering of input columns~\citep{pfefferle_nanotabpfn_2025}. Hence, when a feature is randomly designated as the sensitive attribute during pretraining, its index in the table does not have an impact on predictions.} 

\rev{To discourage the learned representation from encoding sensitive information, the secondary prediction head that predicts $\hat{s}_{\text{qy}}$ is preceded by a gradient reversal layer (GRL)~\cite{ganin2015grl}. The head itself is trained to minimize prediction error of $s_{\text{qy}}$, and would naturally exploit any information about $s$ encoded in the transformer's representation. However, the GRL multiplies the gradient signal passing to the transformer by $-1$, such that the representation is adversarially trained to contain no information about $s$, making the prediction head's task harder.} Intuitively, the GRL forces the transformer to reduce reliance on rows and features that are informative for $s$, \rev{including proxies for $s$, leading to fairer predictions when the same representation is used by the $y$ head.}

A key advantage of the GRL formulation in our setting is that it does not require training a separately optimized adversary for each sampled fairness task, even though the designated sensitive attribute changes across tasks during pretraining. In contrast, an adversarial setup with separately trained classifier adversaries would require fitting or re-optimizing a new adversary for each sampled fairness task during pretraining. Such per-task adversarial training would substantially increase pretraining complexity and make large-scale task sampling less practical. 

\rev{Since the fairness tasks used for pretraining sample the sensitive attribute at random from the available features (Section~\ref{sec:fairness_task_sampling}), one may worry that the GRL degrades the embedding broadly rather than removing only sensitive information. The key design point is that the sensitive attribute is removed from $X'$ and provided through a dedicated encoder as an explicit token for every context row (Algorithm~\ref{alg:fairtfm-training}, lines 3 and 8--9). The $s$-head only predicts $s$, so the reversed gradient only penalizes information about the attribute signalled by the in-context $s$ tokens; the learned invariance to $s$ is conditional and in-context. Moreover, since each feature of $X'$ is treated as sensitive in some sampled tasks and not in others, the overall objective is not optimized by degrading any fixed feature. In the running example, the query representations suppress gender \emph{and its proxies} because the bank designates gender as the sensitive attribute in the context examples.}

At inference time, we do not assume access to the sensitive attribute $s_\mathrm{qy}$. Since the transformer is trained to create representations that do not contain information about $s_\mathrm{qy}$, providing it would not be particularly fruitful. Hence, we replace $s_\mathrm{qy}$ values with a fixed learnable parameter during both pretraining and inference. This allows the transformer encoder to operate on $d$-dimensional tokens representing $(x_i,y_i,s_i)$ and produce query sample embeddings, while ensuring that $s_\mathrm{qy}$ is never directly exposed to the model. Beyond obviating the need for direct access to sensitive information at inference, this design discourages the model from exploiting $s_\mathrm{qy}$ as a predictive shortcut, which would undermine the objective of the GRL.

\subsection{Training objective}\label{sec:training_objective}
\begin{algorithm}[t]
\caption{\ftfm pretraining}
\label{alg:fairtfm-training}
\begin{algorithmic}[1]
\Require Data prior $p(\mathcal{D})$, \ftfm{} parameters $\theta$, fairness weight $\lambda$
\While{not converged}
    \State Sample a dataset $(X',y) \sim p(\mathcal{D})$
    \State Produce $X$ by randomly removing one feature from $X'$ as the sensitive attribute $s$
    \If{$s$ is continuous}
        \State Discretize $s$ into categorical groups via ``stick-breaking'' quantile bins
    \EndIf
    \State Split $(X,y,s)$ into context $(X_{\mathrm{ctx}},y_{\mathrm{ctx}},s_{\mathrm{ctx}})$ and query $(X_{\mathrm{qy}},y_{\mathrm{qy}},s_{\mathrm{qy}})$
    \State Mask $y_{\mathrm{qy}}$ and replace $s_{\mathrm{qy}}$ with the learnable sensitive attribute parameter
    \State Encode $X_i$, $y_i$, and $s_i$ into $d$-dimensional vectors using their respective MLP encoders
    \State Encode the $d$-dimensional triplets $(X_i,y_i,s_i)$ with the shared transformer encoder
    \State Predict query labels $\hat{y}_{\mathrm{qy}}$ with the main head
    \State Predict query sensitive attributes $\hat{s}_{\mathrm{qy}}$ with the adversarial head through a GRL
    \State Compute $\mathcal{L} = \mathrm{CE}(\hat{y}_{\mathrm{qy}}, y_{\mathrm{qy}}) + \lambda \, \mathrm{CE}(\hat{s}_{\mathrm{qy}}, s_{\mathrm{qy}})$
    \State Update $\theta$ by backpropagation; the GRL reverses gradients from the sensitive-attribute head into the encoder
\EndWhile
\end{algorithmic}
\end{algorithm}

\rev{\ftfm is trained from scratch on purely synthetic fairness tasks derived from TabICL's prior \cite{qu2025tabicl} (Appendix~\ref{app:hyperparam}).} The transformer, shown in Figure~\ref{fig:overview:sub2}, produces an embedding that captures row- and feature-wise interactions among the context instances, and the two MLP heads output logits for the query label ($\hat{y}_{\text{qy}}$) and corresponding sensitive attribute ($\hat{s}_{\text{qy}}$). We optimize the model using the following joint cross-entropy objective:
\begin{equation}
    \mathcal{L} = \text{CE}(\hat{y}_{\text{qy}}, y_{\text{qy}}) + \lambda \cdot \text{CE}(\hat{s}_{\text{qy}}, s_{\text{qy}}).
    \label{eq:main_loss}
\end{equation} 

Minimizing $\mathcal{L}$ encourages accurate prediction of both the target label $y$ and the sensitive attribute $s$, while the GRL before the $s$ head ensures the encoder learns features that are not predictive of $s$. 
This objective naturally induces a fairness-accuracy trade-off, especially when the target label $y$ and sensitive attribute $s$ are correlated. The parameter $\lambda$ in Eq.~\ref{eq:main_loss} controls the strength of this trade-off. As summarized in Algorithm~\ref{alg:fairtfm-training}, each pretraining step instantiates a new fairness task, forms context and query sets for ICL, masks the query label and sensitive attribute to preserve the intended inference setting, and then updates the shared transformer encoder and other components through the coupled label-prediction and sensitive-attribute losses.

\section{Results}
\label{sec:results}
In this section, we describe the experimental setup, including the evaluation tasks, baseline models, and metrics, before discussing the main empirical results.  To facilitate reproducibility, we provide the inference code, pretrained checkpoints, and scripts for reproducing
the main results at \href{https://github.com/patrikken/FairTFM-inference}{\texttt{github.com/patrikken/FairTFM-inference}}.
 
\subsection{Experimental setup}

\pgraph{Datasets} We evaluate our model on 120 fairness tasks derived from the 2018 1-Year American Community Surveys~\citep{ding2021retiring}, accessed through the \texttt{folktables} library, which is released under the MIT License.\footnote{https://github.com/socialfoundations/folktables} We generate these tasks by varying three factors: (i) the prediction problem (ACSIncome, ACSPublicCoverage, ACSMobility, ACSEmployment, and ACSTravelTime), (ii) the sensitive attribute (Gender, Age, and Race), and (iii) the state from which the data is sampled (eight US states). This yields a total of $5 \times 3 \times 8 = 120$ real-world fairness tasks not used during training. Additional details are provided in Appendix~\ref{app:dataset}. \rev{To test generalization beyond a single data source, we further evaluate on 12 fairness tasks built from six widely used non-ACS benchmarks (Appendix~\ref{app:other-datasets}).}

\pgraph{Base models} We compare our method against standard machine learning baselines, including logistic regression (LR), random forest (RF), XGBoost (XGB), and $k$-nearest neighbours (KNN) in their default scikit-learn configuration~\cite{pedregosa2011scikit}. We also include recent TFMs, namely TabPFNv2.5 and TabICLv2. Because these models do not incorporate any fairness intervention, they provide strong baselines for assessing the fairness limitations of existing approaches on our benchmark. These models do not receive $s_\text{qy}$ at inference, similar to \ftfm{}. \rev{We additionally compare against FairPFN~\citep{robertsonfairpfn}, the only prior TFM that incorporates fairness during pretraining. FairPFN targets counterfactual fairness rather than the statistical notions we optimize, so it is not designed to minimize DP, EOD, or EOP; we include it as the closest existing approach to ours, to test whether pretraining for a causal fairness notion also yields group-fairness benefits. We use the publicly released FairPFN checkpoint, so its backbone size and pretraining prior differ from ours; this comparison should therefore be read as a comparison between published methods rather than as a controlled ablation of the fairness objective.}

\pgraph{Metrics} In addition to accuracy and AUCROC, we report three fairness metrics: demographic parity difference (DP), equalized odds difference (EOD), and equal opportunity difference (EOP). Definitions and implementation details are provided in Appendix~\ref{app:fairness-metrics}. Unless otherwise noted, we report metric values averaged across all 120 tasks to summarize overall fairness--accuracy trends. 

\subsection{Experimental Results}\label{sec:results_subsec}
In this section, we present a comprehensive empirical evaluation of \ftfm across three complementary settings. First, we benchmark its performance in terms of both accuracy and fairness against classical methods and recent tabular foundation models. Second, we compare against fairness-aware variants of classical approaches to assess trade-offs under explicit fairness constraints. Third, we evaluate \ftfm as a fairness-enhancing representation learner, examining its ability to produce embeddings that support fairer downstream predictions.   

 \begin{figure*}
    \centering
    \includegraphics[width=.95\linewidth]{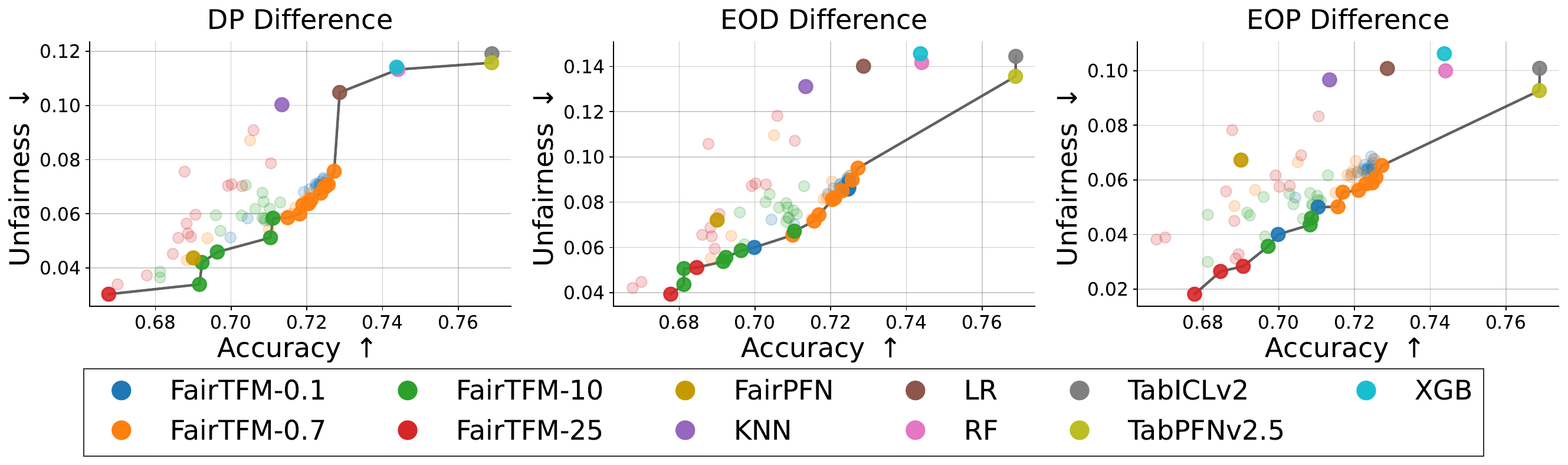}
    \caption{Pareto front between accuracy and fairness for various models. $\uparrow$ indicates higher is better (accuracy) and $\downarrow$ indicates smaller is better (unfairness).}
    \label{fig:main-results}
        \vspace{-10pt}
\end{figure*}
\pgraph{Fairness-accuracy trade-off across tasks} Figure~\ref{fig:main-results} shows our main results: the Pareto front between accuracy and fairness for various models. 
For \ftfm, we pretrain variants with $\lambda \in \{0.1, 0.7, 10, 25\}$ to control the fairness--accuracy trade-off, and display that trade-off for several pretraining checkpoints to more fully trace out the frontier. TabICLv2 and TabPFNv2.5 achieve the highest average accuracy across tasks, but they also occupy the most unfair region of the Pareto frontier, suggesting that their predictive gains come with a substantial fairness cost. In contrast, the classical machine learning baselines are generally less accurate, yet they remain similarly unfair.

\ftfm consistently improves fairness metrics, and for a range of $\lambda$ parameters maintains competitive---and often stronger---accuracy compared to classical baselines such as LR and KNN. Table~\ref{tab:baseline_comparison} in Appendix~\ref{app:aggregated_results} complements Figure~\ref{fig:main-results} by reporting the corresponding aggregated metrics over all 120 tasks, including standard deviations across three random seeds. Whereas Figure~\ref{fig:main-results} emphasizes the Pareto frontier traced out by different checkpoints and fairness weights, Table~\ref{tab:baseline_comparison} makes the overall pattern explicit at the level of final checkpoints: the strongest unconstrained TFMs achieve the highest accuracy but are also the least fair, while \ftfm variants deliver substantial reductions in DP, EOD, and EOP with only a moderate loss in predictive performance. \rev{Table~\ref{tab:percent-improvement} in Appendix~\ref{app:percent-improvement} reports the relative changes against the strongest TFM (TabPFNv2.5), the strongest classical model (XGB), and LR: \ftfm improves fairness by 32--75\% for accuracy costs of 2--11\%, and even dominates LR on all metrics at $\lambda=0.7$. Tables~\ref{tab:baseline_comparison} and~\ref{tab:percent-improvement} report the last pretraining checkpoint of each \ftfm variant, while the Pareto-dominant checkpoints visible in Figures~\ref{fig:main-results} and~\ref{fig:result-other-dataset} yield better trade-offs, but would require model selection on a validation set.}

The results also highlight the role of $\lambda$ in shaping the accuracy-fairness trade-off. Larger values of $\lambda$ (e.g., $\lambda = 25$) move the model toward the fairer region of the Pareto frontier at the expense of predictive accuracy. Smaller values of $\lambda$, by contrast, prioritize accuracy and therefore place the model in less fair regions of the frontier. The Pareto front ends up tracing a nearly linear path through accuracy-unfairness space, with \ftfm filling out a large segment. These trends indicate that our training objective provides a simple and effective mechanism for navigating different operating points depending on the fairness requirements of the application. We observe the same trend when measuring predictive performance using AUCROC, as shown in Figure~\ref{fig:results-auc-fairness} \rev{of Appendix~\ref{app:additional-results}}.

\rev{FairPFN behaves differently from the unconstrained baselines: it is fairer than all of them on all three metrics, and in Table~\ref{tab:baseline_comparison} attains the lowest DP of any final checkpoint (0.043). Pretraining for a causal fairness notion therefore does transfer part of its benefit to statistical group fairness, supporting our broader claim that fairness can be acquired during pretraining and translate to ICL predictions. However, for FairPFN this comes at a large predictive cost; FairPFN has the lowest AUCROC of all models we evaluate here (0.695). In every panel of Figures~\ref{fig:main-results} and~\ref{fig:results-auc-fairness} \ftfm provides operating points that Pareto-dominate it, being simultaneously fairer and more predictive. Just as important, FairPFN exposes no mechanism for navigating the fairness--utility trade-off: it yields a single operating point, whereas $\lambda$ lets a practitioner choose one that matches the application. Part of the predictive gap is expected, since FairPFN optimizes fair outcomes under interventions on structural causal models rather than accuracy on real labels; the comparison is thus evidence that the two fairness notions are not interchangeable, not that FairPFN fails at its own objective.}

\rev{
\begin{figure*}[t]
    \centering
    \includegraphics[width=.95\linewidth]{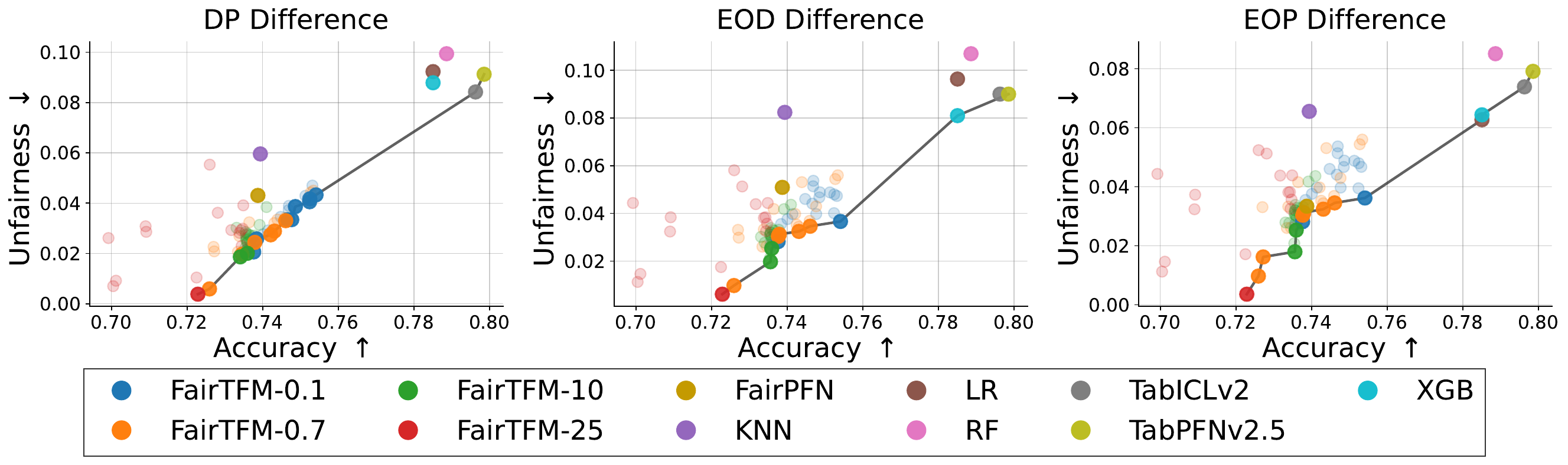}
    \caption{Pareto front between accuracy and fairness for various models on the 12 fairness tasks beyond ACS PUMS (Appendix~\ref{app:other-datasets}). $\uparrow$ indicates higher is better (accuracy) and $\downarrow$ indicates smaller is better (unfairness).}
    \label{fig:result-other-dataset}
    \vspace{-10pt}
\end{figure*}
\pgraph{Generalization beyond ACS PUMS tasks} Figure~\ref{fig:result-other-dataset} shows the Pareto front on the 12 non-ACS tasks. The conclusions match those from Figure~\ref{fig:main-results}: TabICLv2 and TabPFNv2.5 are the most accurate but occupy the most unfair region, while \ftfm variants trace the low-unfairness end of the front, with $\lambda$ controlling the trade-off. The AUCROC view (Figure~\ref{fig:result-other-dataset-auc} in Appendix~\ref{app:other-datasets}) further shows that \ftfm improves fairness without a commensurate loss in ranking performance. FairPFN behaves as on the ACS benchmark: it is fairer than every unconstrained baseline on all three metrics, but at a predictive cost that again leaves it Pareto-dominated by \ftfm{}, under both accuracy and AUCROC. Table~\ref{tab:other-datasets} in Appendix~\ref{app:other-datasets} reports the corresponding aggregated metrics and demonstrates the gap directly. At essentially the same accuracy (0.734 vs.\ 0.737), \ftfm{}-10 is fairer than FairPFN on all three metrics and 7.5 points higher in AUCROC.}

\begin{figure}[t]
    \centering
    \includegraphics[width=\linewidth]{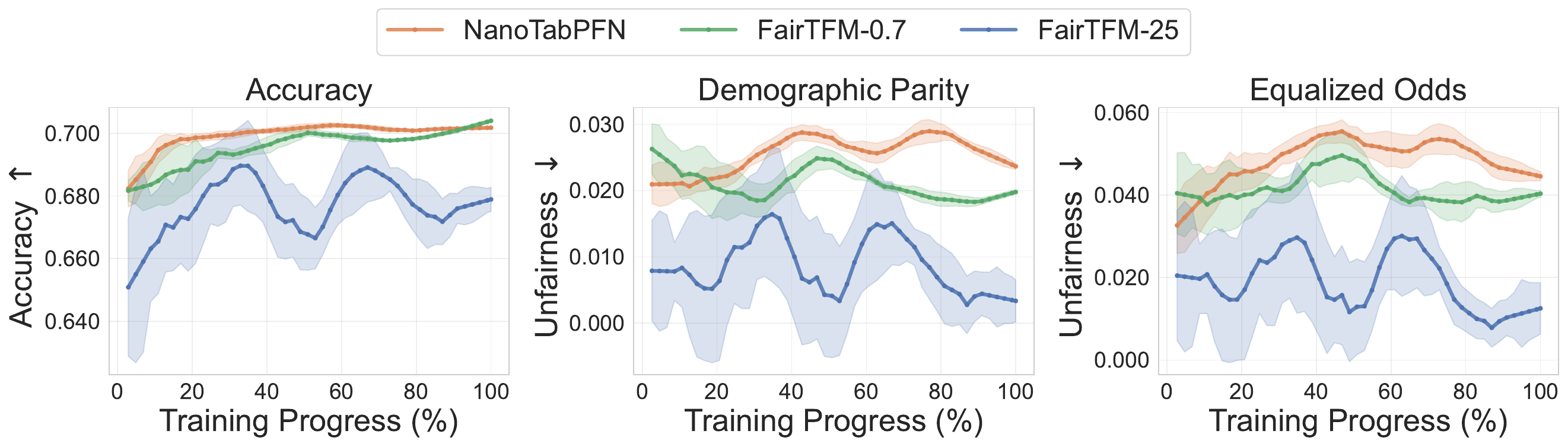}
    \caption{Training dynamics comparison between nanoTabPFN (without fairness constraint) and \ftfm{} with $\lambda \in \{0.7, 25\}$. }
    \label{fig:dynamics-during-training}
        \vspace{-10pt}
\end{figure}

\pgraph{Performance dynamics during training}
Figure~\ref{fig:dynamics-during-training} provides a finer-grained view of how fairness emerges over the course of pretraining, tracking accuracy and fairness (DP and EOD) for \ftfm with $\lambda \in \{0.7, 25\}$ and for nanoTabPFN across checkpoints. Both \ftfm{} variants achieve a fairness advantage over nanoTabPFN very early in training and retain this advantage throughout optimization. \rev{Since both models share the same data prior, backbone, and training budget, differing only in the fairness-specific components, this comparison isolates the effect of our intervention (full discussion in Appendix~\ref{app:training-dynamics}).}

\pgraph{Comparing \ftfm against classical fairness-aware models}
The results above show that \ftfm improves fairness substantially while remaining competitive with standard classical baselines such as LR and KNN. A stronger comparison, however, is against classical methods that are themselves explicitly optimized for fairness. To this end, we evaluate three tabular models---LR, RF, and XGB---augmented with the Exponentiated Gradient (EG) reduction of~\citet{agarwal2018reductions}, which enforces group-fairness constraints during training. We focus on ACSIncome in Alabama (AL) and instantiate three fairness tasks by varying the sensitive attribute over gender, race, and age. For each task, we use a random 80/20 train--test split and sweep the EG fairness-violation tolerance over $[0.01, 0.02, \ldots, 0.1, 0.2, \ldots, 1.0]$, with finer resolution in the low-violation regime to better characterize the high-fairness end of the trade-off. We average the results across three random seeds. This setup allows us to compare \ftfm not only to strong predictive baselines, but also to established in-processing fairness methods under a matched downstream evaluation protocol.

\begin{figure}[t]
    \centering
    \includegraphics[width=.95\linewidth]{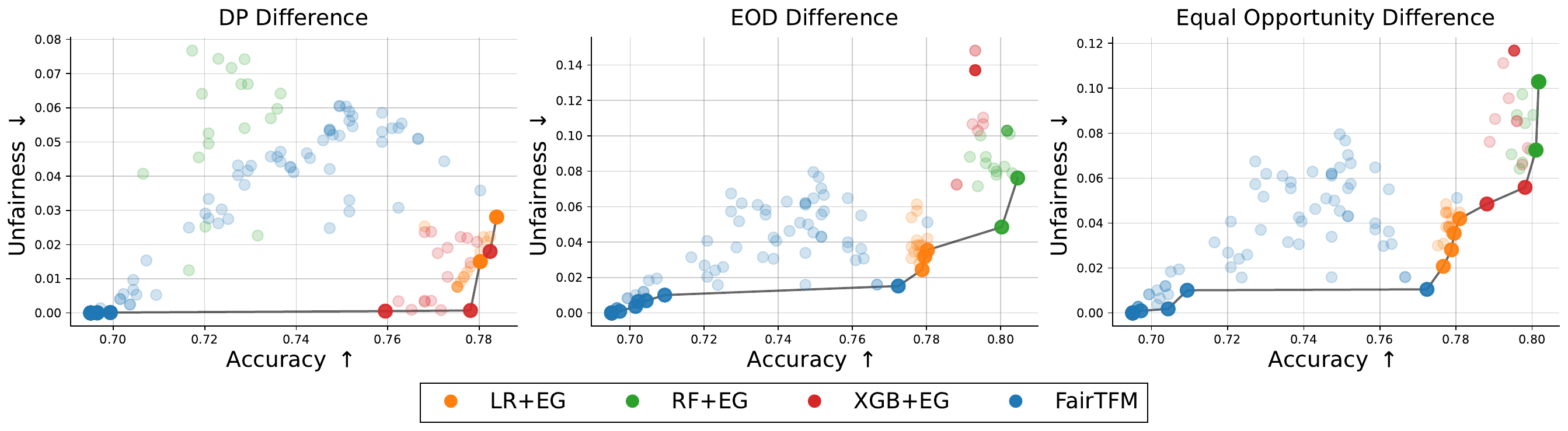}
   \caption{Fairness-accuracy trade-off comparison with task-specific classical models trained with fairness constraints using Exponentiated Gradient (EG), on ACSIncome (Alabama) with gender as the sensitive attribute. \rev{Results for race and age are shown in Figure~\ref{fig:exgrad-baseline-race-age} of Appendix~\ref{app:additional-results}.}}
    \label{fig:exgrad-baseline}
        \vspace{-10pt}
\end{figure}

Figure~\ref{fig:exgrad-baseline} shows that \ftfm remains highly competitive even against these fairness-constrained baselines. Across all three sensitive attributes, its checkpoints trace out a broad Pareto frontier, indicating that a single pretrained model can realize multiple fairness--accuracy operating points without retraining. For gender, \ftfm attains especially strong trade-offs for EOD and EOP, matching or improving upon the frontier formed by EG-based baselines while remaining competitive on DP. For race \rev{(Figure~\ref{fig:exgrad-baseline-race-age} in Appendix~\ref{app:additional-results})}, where all methods incur larger fairness gaps, \ftfm{} still spans a wide and competitive portion of the frontier, particularly at the lower-unfairness end. For age, the comparison is more metric-dependent: EG-based classical models achieve stronger accuracy--fairness trade-offs for EOD and EOP, whereas \ftfm remains competitive on DP but exhibits a clearer trade-off between predictive performance and these stricter parity criteria. Overall, the main advantage of \ftfm{} is not that it dominates every baseline on every metric, but that it delivers competitive Pareto-efficient solutions across heterogeneous fairness tasks in a single forward pass, whereas the classical alternatives must be retrained with task-specific fairness constraints for each new setting. Moreover, as shown in Figure~\ref{fig:exgrad-baseline-auc} of Appendix~\ref{app:additional-results}, \ftfm provides the best overall Pareto front when predictive performance is measured through AUCROC.

\rev{We also evaluated the EG baselines on the 12 non-ACS tasks with results in Appendix~\ref{app:eg-other-datasets}. Each \ftfm setting improves the metrics that are relevant to each EG variant---by 20--33\% for \ftfm{}-0.7 and 79--82\% for \ftfm{}-25---while achieving substantially higher AUCROC: EG reduces XGBoost's AUCROC from 0.803 to about 0.70, whereas \ftfm{}-0.7 remains at 0.803.}

\pgraph{\ftfm as a fair representation learner}
We hypothesize that the \ftfm encoder produces representations that are useful for downstream fair prediction. \rev{Fair representation learning methods learn a map $g$ of the input that suppresses information about the sensitive attribute while preserving task-relevant signal; formally:}

\begin{definition}[Fair Representation]
\label{def:fair-representation}
Let $(X, Y, S)$ be random variables where
$X \in \mathbb{R}^{m \times d}$ is the input,
$Y \in \mathcal{Y}$ is the target label, and
$S \in \mathcal{S}$ is a sensitive attribute. Let $g: \mathbb{R}^{m \times d} \to \mathbb{R}^{m \times d'}$ be a representation map and define $Z := g(X)$.
We say that $Z$ is a fair representation with respect to $S$ if
$Z \perp\!\!\!\perp S \mid Y$.
\end{definition}
\rev{Definition~\ref{def:fair-representation} is conceptually aligned with the objective of \ftfm, which encourages the learned query embedding to suppress sensitive-attribute information while preserving task-relevant signal. We compare \ftfm{} embeddings against two task-specific pre-processing methods from the \texttt{fairlearn} library~\citep{bird2020fairlearn}: Correlation Remover (CR) and Learning Fair Representations (LFR)~\citep{zemel2013learning}, on ACSIncome (Alabama) with gender, race, and age as sensitive attributes. Notably, \ftfm{} embeddings are computed in a single forward pass without any task-specific gradient updates. For each type of representation as well as the raw features, we train downstream classifiers (LR, RF, and XGBoost), and compare their accuracy and fairness properties. Full protocol details are in Appendix~\ref{app:fair-rep-details}.}

\rev{
\begin{figure}[t]
    \centering
    \includegraphics[width=.95\linewidth]{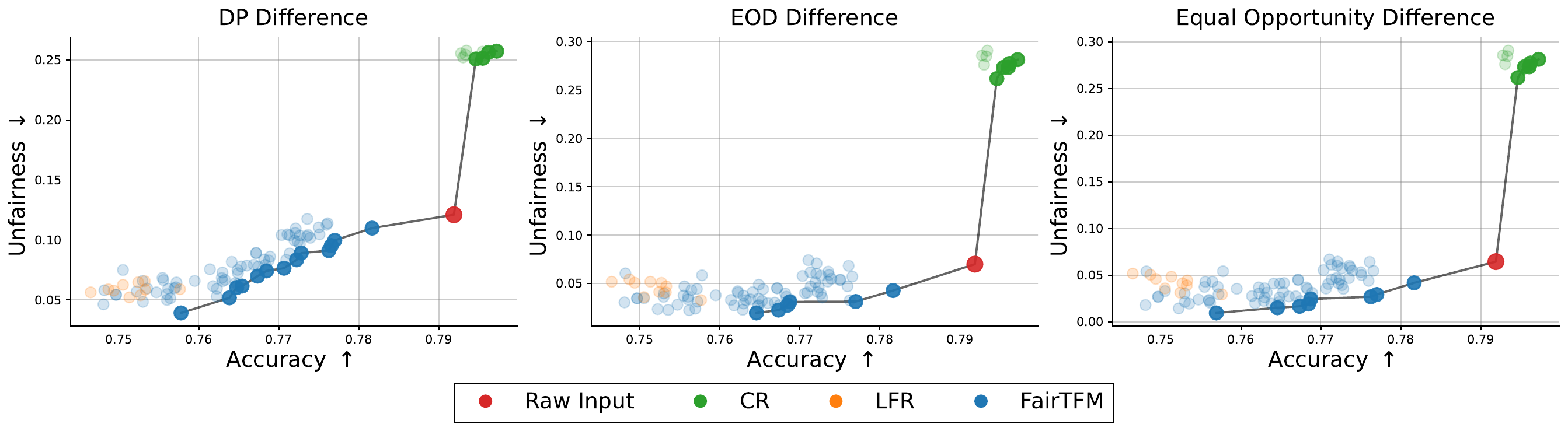}
   \caption{Pareto front of XGBoost models trained with different data representations, with gender as the sensitive attribute. Results for race and age, and for RF and LR as downstream models, are shown in Appendix~\ref{app:fair-rep-details}.}
    \label{fig:fair-rep-XGB-gender}
\end{figure}
}

 Figure~\ref{fig:fair-rep-XGB-gender} showcases the Pareto front of XGBoost models trained using data representation from different fair methods\rev{, with gender as the sensitive attribute}. As can be seen, \ftfm can provide competitive Pareto points on downstream tasks compared to raw data and dominates LFR, whereas CR severely worsens fairness. The results show the ability of \ftfm to generate embeddings that encode less information about the given sensitive attributes while still being predictive of the target. \ftfm{} stands out due to its ability to adapt to new tasks via ICL without needing task-specific optimization. \rev{The remaining sensitive attributes (race and age) and downstream models (RF and LR) in Appendix~\ref{app:fair-rep-details} show a similar trend.}

\rev{
\begin{table}[t]
\centering
\caption{Linear probe AUC for predicting the sensitive attribute $s$ and the label $y$ on different representations. We show AUC mean $\pm$ std over three seeds and the three sensitive attributes on ACSIncome, Alabama. \ftfm displays a controllable trade-off of reducing the predictability of the sensitive attribute with some obfuscation of the label.}
\label{tab:probing}
\renewcommand{\arraystretch}{1.1}
\setlength{\tabcolsep}{6pt}
\begin{tabular}{lcccc}
\toprule
\textbf{Probe target (AUC)} & Raw input & nanoTabPFN & \ftfm{}-0.7 & \ftfm{}-25 \\
\midrule
Sensitive attribute $s$ & 0.783 $\pm$ 0.11 & 0.746 $\pm$ 0.10 & 0.646 $\pm$ 0.14 & 0.542 $\pm$ 0.11 \\
Label $y$ & 0.834 $\pm$ 0.02 & 0.823 $\pm$ 0.03 & 0.805 $\pm$ 0.04 & 0.787 $\pm$ 0.03 \\
\bottomrule
\end{tabular}
\end{table}
\pgraph{Probing the learned representations} To verify that the GRL removes information about the designated sensitive attribute rather than degrading the embedding broadly (Section~\ref{sec:method}), we train linear probes to predict (a) the sensitive attribute $s$ and (b) the label $y$ from three representations: the raw input features $X$, the query embeddings of nanoTabPFN, and the query embeddings of \ftfm. For the embedding-based representations, the query embeddings of the initial test set are treated as a new dataset for each probing task. For both sensitive-attribute and label prediction, this dataset is randomly split into 80\% training and 20\% testing subsets. The probing model (XGBoost) is trained on the training split, and performance is evaluated on the held-out test split using AUC.

Table~\ref{tab:probing} shows three things. First, the nanoTabPFN embedding retains most of the sensitive information present in the raw features (AUC 0.746 vs.\ 0.783) even though $s_\text{qy}$ itself is not an input. Second, \ftfm substantially reduces the sensitive-attribute AUC, approaching random guessing (0.542 at $\lambda=25$). Third, task-relevant information is largely preserved (AUC 0.834 $\to$ 0.805 at $\lambda=0.7$). These probes support the idea that suppression of sensitive information is attributable to the adversarial objective and is selective.}

\section{Conclusion}
\looseness=-1
We present \ftfm, the first pretraining framework for tabular foundation models that provides group fairness while preserving the single-pass inference setting of in-context learning. The key idea is to synthesize fairness tasks during pretraining and couple label prediction with adversarial sensitive-attribute prediction through a gradient reversal layer. On 120 ACS-based fairness tasks \rev{and 12 additional tasks from six non-ACS benchmarks}, the resulting models consistently improve demographic parity, equal opportunity, and equalized odds while maintaining competitive accuracy. Beyond the aggregate benchmark results, we showed that these gains emerge throughout training, remain competitive against fairness-constrained classical baselines, and transfer to downstream fair-representation settings\rev{, with probing experiments confirming that the learned embeddings selectively remove sensitive information}. Taken together, these results suggest that fairness can be incorporated directly into the pretraining of TFMs, rather than introduced only through task-specific post-hoc correction as done in prior work.

\paragraph{Limitations and future work} \looseness=-1 Despite the strong empirical performance of \ftfm{}, our study has several limitations. First, our pretraining strategy relies only on TabICL's prior generation, which provides scale and diversity but may not capture all of the semantic and societal structure of real sensitive attributes in downstream deployments. Second, while \ftfm{} is broadly competitive, the results also show that it does not dominate specialized fairness-aware baselines on every metric or every task configuration, especially in the stricter settings where age is used as sensitive attribute. We view these limitations not as drawbacks of the overall approach, but as evidence that fairness-aware pretraining opens a rich new research direction. In particular, our findings pave the way for future work on better fairness-task priors, improved trade-off control during pretraining, and foundation models that can adapt their fairness behaviour more explicitly to downstream deployment requirements. Finally, we emphasize that our experiments rely on nanoTabPFN, a lightweight backbone chosen to isolate the effect of fairness-aware pretraining. We believe the proposed framework is largely orthogonal to architectural scaling, and integrating it into larger state-of-the-art TFMs such as TabPFNv2.5 or TabICLv2 may yield substantially stronger fairness–accuracy trade-offs through improved representation learning and richer pretrained priors.

\bibliography{bib.bib} 
\bibliographystyle{abbrvnat}

\clearpage
\appendix

\section{Additional Related Work}
\label{app:related-work}

\pgraph{Fairness interventions} Classical fairness-aware learning methods are often grouped by where the intervention occurs in the pipeline~\cite{cresswell2025trustworthy}. Pre-processing methods attempt to reduce bias directly in the data, for example through label massaging~\citep{kamiran2009classifying} or fair representation learning~\citep{zemel2013learning}. In-processing methods instead modify the training objective to encourage fairer behavior during optimization~\citep{zhang2018mitigating}. Post-processing methods operate on a trained model's outputs to enforce fairness constraints after training~\citep{hardt2016equality}. These families of methods have been influential, but they typically assume task-specific model access, retraining, or output recalibration.

\pgraph{Fairness in in-context learning} These assumptions become restrictive for foundation models used through ICL, where predictions are generated without task-specific parameter updates. In our setting, the goal is not to retrofit fairness onto a frozen predictor after deployment, but to pretrain a model whose internal representations already support fairer predictions at inference time. Our method therefore combines aspects of in-processing and representation learning: fairness is encouraged during training, yet inference remains a single forward pass without additional optimization.

\pgraph{Tabular foundation models} Work on deep learning for tabular data has evolved from specialized supervised architectures~\citep{gorishniy2021revisiting} to pretrained tabular foundation models that generalize across tasks through ICL. TabPFN~\citep{hollmanntabpfn} established this direction by showing that transformers trained on synthetic supervised tasks can perform strong zero-shot prediction. Subsequent models, including TabPFNv2.5~\citep{grinsztajn2025tabpfn} and TabICLv2~\citep{qu2026tabiclv2}, improved efficiency and scale, while models such as TabDPT~\citep{ma2024tabdpt} and related real-data pretraining approaches~\citep{garg2025real} sought to better align training with real-world tabular distributions.

\pgraph{Fairness-aware TFMs} Among prior TFM work, FairPFN~\citep{robertsonfairpfn} is the closest to our setting because it also incorporates fairness during pretraining. However, FairPFN is designed around a causal notion of fairness, using structurally generated datasets with biased and fair outcomes to target counterfactual fairness~\citep{kusner2017counterfactual}. Our focus is different: we study statistical group fairness notions, namely demographic parity, equal opportunity, and equalized odds, which are more directly aligned with the evaluation metrics commonly used in fair classification benchmarks. Statistical fairness is data-driven and focuses on distributional fairness (outcome-based), while counterfactual fairness is more interventional and does not necessarily ensure outcome parity across groups

\section{Datasets}
\label{app:dataset}
 In this section, we provide more details about the datasets used for evaluation. We describe the prediction tasks and the dataset construction. 
\subsection{Prediction Task Details}
We construct our benchmark from tasks provided by the \texttt{folktables} benchmark~\citep{ding2021retiring}, which is derived from the American Community Survey (ACS) Public Use Microdata Sample (PUMS). In particular, we consider five binary prediction tasks commonly used in prior work for fairness evaluation:

\begin{itemize}[leftmargin=*, nosep]
    \item \textbf{ACSIncome}: predict whether an individual's annual income exceeds \$50,000. Following the standard task definition, we restrict the data to individuals older than 16 who worked at least one hour per week during the previous year and earned at least \$100.

    \item \textbf{ACSMobility}: predict whether an individual lived at the same address one year earlier. We focus on individuals between 18 and 35 years old, which makes the task less imbalanced than in the full population, where most individuals do not move within a year.

    \item \textbf{ACSTravelTime}: predict whether an individual's commute exceeds 20 minutes. The task is defined on employed individuals older than 16, and the 20-minute threshold roughly matches the median commute time in the 2018 ACS PUMS data.

    \item \textbf{ACSEmployment}: predict whether an individual is employed. For this task, we consider individuals between 16 and 90 years old.

    \item \textbf{ACSPublicCoverage}: predict whether an individual receives public health insurance coverage. We restrict the sample to individuals younger than 65 with income below \$30,000, thereby focusing on lower-income individuals who are not eligible for Medicare.
\end{itemize}

\subsection{Task Construction}

Generating new learning signals from randomly selected features in pretraining data is a core concept in self-supervised learning~\cite{sui2024self}. In our work, we specifically construct fairness tasks, which is necessary to scale up pretraining, due to insufficient existing datasets labeled with fairness attributes. For each base prediction task, we instantiate fairness evaluation settings by combining it with three sensitive attributes---Gender, Age, and Race---and with data drawn from eight states: Alabama (AL), California (CA), Hawaii (HI), Indiana (IN), Maine (ME), Michigan (MI), New Mexico (NM), and New York (NY). We choose these states to span a range of bias levels reported in the original \texttt{folktables} study~\citep{ding2021retiring}, so that the benchmark includes settings with meaningfully different fairness profiles rather than a narrow slice of the ACS. This yields a total of $5 \times 3 \times 8 = 120$ tasks. In other words, each prediction problem contributes 24 tasks, each sensitive attribute appears in 40 tasks, and each state contributes 15 tasks to the full benchmark. For the sensitive attributes, Age is binarized using a 25-year-old threshold, and Race is restricted to White and Black Americans. Across all task instantiations, dataset sizes vary from roughly 3.5k to 12k samples, and we use a random 80/20 train--test split.

This evaluation design is important for two reasons. First, varying the prediction problem, sensitive attribute, and state allows us to assess fairness across substantially different label distributions, demographic compositions, and regional contexts, rather than tailoring conclusions to a single task configuration. Second, reporting results over the full Cartesian product reduces the risk that observed fairness improvements are driven by a small number of favorable settings. We therefore view performance aggregated over these 120 tasks as a stronger indicator of whether a method learns fairness-aware behavior that transfers across heterogeneous real-world tabular prediction problems. 

\section{Fairness metrics}
\label{app:fairness-metrics}

In this work, we focus on group fairness criteria that quantify disparities in model behavior across demographic groups. Let $\hat{Y} = f(X)$ denote the binary prediction of a classifier, let $Y \in \{0,1\}$ be the ground-truth label, and let $S \in \{0,1\}$ denote the sensitive attribute. We consider the following three standard fairness notions.

\begin{itemize}[leftmargin=*, nosep]
    \item \textbf{Demographic parity (DP)} requires the rate of positive predictions to be the same across groups~\citep{dwork2012fairness}. Formally,
    \begin{equation}
    \label{eq:dp}
        \mathbb{P}(\hat{Y}=1 \mid S=0) = \mathbb{P}(\hat{Y}=1 \mid S=1).
    \end{equation}

    \item \textbf{Equalized odds (EOD)} requires the predictor to have the same true positive and false positive rates across groups~\citep{hardt2016equality}. Equivalently, for each label value $y \in \{0,1\}$,
    \begin{equation}
    \label{eq:eodd}
        \mathbb{P}(\hat{Y}=1 \mid S=0, Y=y) = \mathbb{P}(\hat{Y}=1 \mid S=1, Y=y).
    \end{equation}

    \item \textbf{Equal opportunity (EOP)} focuses only on parity of true positive rates across groups. It can be viewed as the $y=1$ special case of equalized odds:
    \begin{equation}
    \label{eq:eop}
        \mathbb{P}(\hat{Y}=1 \mid S=0, Y=1) = \mathbb{P}(\hat{Y}=1 \mid S=1, Y=1).
    \end{equation}
\end{itemize}

In the experiments, we report empirical disparity versions of these metrics. For demographic parity, we use the absolute difference in the expected positive prediction rate across groups:

\begin{equation}
\label{eq:d_dp}
    \mathrm{DP} = \left| \underset{x\mid S=0}{\mathbb{E}}\left[\mathbb{I}\{\hat{Y}=1\}\right] - \underset{x\mid S=1}{\mathbb{E}}\left[\mathbb{I}\{\hat{Y}=1\}\right] \right|.
\end{equation}
Where $\mathbb{I}(\cdot)$ denotes the indicator function.

For the equalized-odds-based metrics, we define the group gaps in false positive rate and true positive rate using the same expectation notation:
\begin{equation}
\label{eq:alpha0}
    \delta_{\mathrm{FPR}} = \left| \underset{x\mid S=0, Y=0}{\mathbb{E}}\left[\mathbb{I}\{\hat{Y}=1\}\right] - \underset{x\mid S=1, Y=0}{\mathbb{E}}\left[\mathbb{I}\{\hat{Y}=1\}\right] \right|,
\end{equation}
\begin{equation}
\label{eq:alpha1}
    \delta_{\mathrm{TPR}} = \left| \underset{x\mid S=0, Y=1}{\mathbb{E}}\left[\mathbb{I}\{\hat{Y}=1\}\right] - \underset{x\mid S=1, Y=1}{\mathbb{E}}\left[\mathbb{I}\{\hat{Y}=1\}\right] \right|.
\end{equation}
We then report
\begin{equation}
\label{eq:d_eod}
    \mathrm{EOD} = \max\left(\delta_{\mathrm{FPR}}, \delta_{\mathrm{TPR}}\right)
\end{equation}
\begin{equation}
\label{eq:d_eop}
    \mathrm{EOP} = \delta_{\mathrm{TPR}}.
\end{equation}

Smaller values of $\mathrm{DP}$, $\mathrm{EOD}$, and $\mathrm{EOP}$ indicate fairer behaviour, with zero corresponding to perfect parity under the respective criterion. We use these empirical gaps because they provide an interpretable summary of group-level disparities and are standard in fairness evaluations for binary classification.

\section{Model Architecture and Hyperparameters}
\label{app:hyperparam}
Our model builds on the nanoTabPFN architecture from \texttt{TFM-Playground}~\footnote{https://github.com/automl/TFM-Playground}, which we use as a lightweight transformer backbone for fairness-aware pretraining. Concretely, we use a model with 6 transformer layers, 6 attention heads, embedding dimension 192, and feed-forward hidden dimension 192. Consistent with the main architecture described in Section~3, we augment this backbone with three input encoders---for features, targets, and sensitive attributes---that map their respective inputs into the shared 192-dimensional token space. The label-prediction head and the sensitive-attribute head are both implemented as two-layer MLPs with hidden size 768.

\paragraph{Pretraining setup.}
We pretrain on 300{,}000 synthetically generated tabular datasets sampled from the TabICL's prior implementation~\citep{qu2026tabiclv2}. Each sampled task contains 150 datapoints, 6 features, and 2 classes, and training is performed with batch size 32. During pretraining, the model receives context and query sets as described in Section~3, with the query label masked and the query sensitive attribute replaced by the learned mask token. All reported \ftfm{} results are obtained from checkpoints trained under this shared setup, varying only the fairness weight $\lambda$ in the joint objective.

\paragraph{Optimization details.}
We optimize the model using Schedule-Free AdamW~\citep{defazio2024road,loshchilov2017decoupled} with learning rate $1 \times 10^{-4}$ and no weight decay. This choice provided stable optimization across the fairness weights considered in the main experiments. For the fairness-aware variants, the only task-level hyperparameter we vary is $\lambda$, which directly controls the strength of the adversarial sensitive-attribute objective and thereby the fairness--accuracy trade-off.

\section{Additional Results}\label{app:additional-results}

\subsection{Aggregated results on the 120 ACS tasks}\label{app:aggregated_results}

\begin{figure}[h]
    \centering
    \includegraphics[width=.95\linewidth]{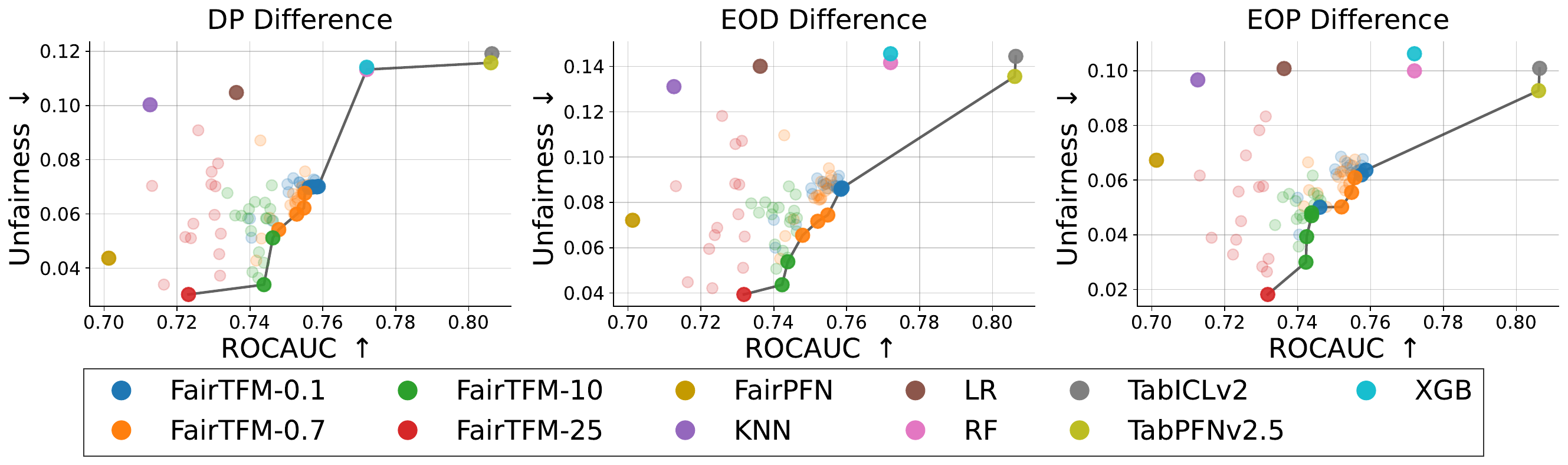}
    \caption{Pareto front between AUCROC and fairness for various models. $\uparrow$ indicates higher is better (accuracy) and $\downarrow$ indicates smaller is better (unfairness).}
    \label{fig:results-auc-fairness}
\end{figure}

\rev{Table~\ref{tab:baseline_comparison} reports the aggregated accuracy and fairness metrics behind Figure~\ref{fig:main-results}, and Figure~\ref{fig:results-auc-fairness} shows the corresponding Pareto front when predictive performance is measured with AUCROC instead of accuracy. The two views agree: unconstrained TFMs sit in the most accurate but least fair region, while \ftfm variants trace the fair end of the front.}

\begin{table}[h]
\centering
\caption{Average accuracy, AUCROC, and fairness metrics across the 120 benchmark tasks, reported as mean $\pm$ standard deviation over three random seeds. 
This table complements Figure~\ref{fig:main-results} by summarizing the final-checkpoint performance of \ftfm{} models. Recent unconstrained TFMs attain the highest average accuracy but also the largest fairness gaps, classical baselines are generally less accurate without being substantially fairer, and \ftfm variants provide better fairness--accuracy trade-offs, with larger $\lambda$ yielding progressively lower unfairness at the cost of reduced accuracy.}
\label{tab:baseline_comparison}
\renewcommand{\arraystretch}{1.0}
\setlength{\tabcolsep}{4pt}
\begin{tabular}{lccccc}
\toprule
\small
\textbf{Model} & \textbf{Accuracy} & \textbf{AUCROC} & \textbf{DP Diff} & \textbf{EOD Diff} & \textbf{EOP Diff} \\
\midrule
TabICLv2   & $\mathbf{0.771 \pm 0.05}$ & \rev{$\mathbf{0.802 \pm 0.07}$} & $0.120 \pm 0.11$ & $0.140 \pm 0.10$ & $0.096 \pm 0.09$ \\
TabPFNv2.5 & $0.770 \pm 0.05$ & \rev{$0.801 \pm 0.07$} & $0.121 \pm 0.12$ & $0.145 \pm 0.11$ & $0.098 \pm 0.10$ \\
\midrule
XGB        & $0.744 \pm 0.06$ & \rev{$0.760 \pm 0.09$} & $0.116 \pm 0.11$ & $0.144 \pm 0.11$ & $0.097 \pm 0.10$ \\
KNN        & $0.716 \pm 0.07$ & \rev{$0.707 \pm 0.10$} & $0.106 \pm 0.11$ & $0.139 \pm 0.10$ & $0.093 \pm 0.09$ \\
RF         & $0.745 \pm 0.06$ & \rev{$0.767 \pm 0.08$} & $0.113 \pm 0.11$ & $0.147 \pm 0.11$ & $0.100 \pm 0.10$ \\
LR         & $0.728 \pm 0.06$ & \rev{$0.733 \pm 0.10$} & $0.107 \pm 0.12$ & $0.134 \pm 0.11$ & $0.094 \pm 0.09$ \\
\midrule
\rev{FairPFN} & \rev{$0.690 \pm 0.07$} & \rev{$0.695 \pm 0.11$} & \rev{$\mathbf{0.043 \pm 0.07}$} & \rev{$0.078 \pm 0.11$} & \rev{$0.069 \pm 0.10$} \\
\midrule
\ftfm{-0.7}   & $0.729 \pm 0.07$ & \rev{$0.754 \pm 0.10$} & $0.077 \pm 0.13$ & $0.098 \pm 0.13$ & $0.065 \pm 0.10$ \\
\ftfm{-1.0}   & $0.726 \pm 0.07$ & \rev{$0.751 \pm 0.10$} & $0.075 \pm 0.12$ & $0.091 \pm 0.12$ & $0.064 \pm 0.09$ \\
\ftfm{-10}   & $0.711 \pm 0.07$ & \rev{$0.743 \pm 0.10$} & $0.063 \pm 0.12$ & $0.076 \pm 0.12$ & $0.051 \pm 0.08$ \\
\ftfm{-25}    & $0.688 \pm 0.06$ & \rev{$0.731 \pm 0.10$} & $0.059 \pm 0.14$ & $\mathbf{0.072 \pm 0.15}$ & $\mathbf{0.025 \pm 0.06}$ \\
\bottomrule
\end{tabular}
\end{table}

\rev{
\subsection{Percent improvement over the strongest baselines}\label{app:percent-improvement}

To make the size of the fairness gains explicit, Table~\ref{tab:percent-improvement} reports the relative change of each \ftfm variant against the strongest TFM baseline (TabPFNv2.5), the strongest classical baseline (XGB), and LR, computed from the averages in Table~\ref{tab:baseline_comparison}. Positive fairness numbers mean fairer (lower DP, EOD, or EOP), and negative accuracy and AUCROC numbers mean lower predictive performance. Three points stand out. First, the fairness gains are large relative to the predictive costs: \ftfm{}-0.7 improves DP, EOD, and EOP by 32--36\% for a 5.3\% accuracy cost against TabPFNv2.5, and only a 2\% accuracy and 0.8\% AUCROC cost against XGB. Second, against LR---the classical baseline closest in accuracy---\ftfm{}-0.7 improves every metric at once: slightly higher accuracy (+0.1\%), higher AUCROC (+2.9\%), and 27--31\% better fairness. Third, the gains grow monotonically with $\lambda$, up to 51--75\% for \ftfm{}-25, confirming that $\lambda$ provides direct control over the fairness--accuracy trade-off.

\begin{table}[h]
\centering 
\caption{Fairness improvement (positive numbers mean fairer) and accuracy and AUCROC change of each \ftfm variant relative to (a) the best TFM (TabPFNv2.5), (b) the best classical model (XGB), and (c) LR, averaged over the 120 benchmark tasks and derived from Table~\ref{tab:baseline_comparison}.}
\label{tab:percent-improvement}
\renewcommand{\arraystretch}{1.1}
\setlength{\tabcolsep}{6pt}
\begin{subtable}{\textwidth}
\centering
\caption{vs.\ TabPFNv2.5}
\begin{tabular}{lccccc}
\toprule
\textbf{Model} & Acc & AUCROC & DP & EOD & EOP \\
\midrule
\ftfm{}-0.7 & $-5.3\%$ & $-5.9\%$ & $\mathbf{+36.4\%}$ & $\mathbf{+32.4\%}$ & $\mathbf{+33.7\%}$ \\
\ftfm{}-1.0 & $-5.7\%$ & $-6.2\%$ & $\mathbf{+38.0\%}$ & $\mathbf{+37.2\%}$ & $\mathbf{+34.7\%}$ \\
\ftfm{}-10 & $-7.7\%$ & $-7.2\%$ & $\mathbf{+47.9\%}$ & $\mathbf{+47.6\%}$ & $\mathbf{+48.0\%}$ \\
\ftfm{}-25 & $-10.6\%$ & $-8.7\%$ & $\mathbf{+51.2\%}$ & $\mathbf{+50.3\%}$ & $\mathbf{+74.5\%}$ \\
\bottomrule
\end{tabular}
\end{subtable}

\vspace{6pt}
\begin{subtable}{\textwidth}
\centering
\caption{vs.\ XGB}
\begin{tabular}{lccccc}
\toprule
\textbf{Model} & Acc & AUCROC & DP & EOD & EOP \\
\midrule
\ftfm{}-0.7 & $-2.0\%$ & $-0.8\%$ & $\mathbf{+33.6\%}$ & $\mathbf{+31.9\%}$ & $\mathbf{+33.0\%}$ \\
\ftfm{}-1.0 & $-2.4\%$ & $-1.2\%$ & $\mathbf{+35.3\%}$ & $\mathbf{+36.8\%}$ & $\mathbf{+34.0\%}$ \\
\ftfm{}-10 & $-4.4\%$ & $-2.2\%$ & $\mathbf{+45.7\%}$ & $\mathbf{+47.2\%}$ & $\mathbf{+47.4\%}$ \\
\ftfm{}-25 & $-7.5\%$ & $-3.8\%$ & $\mathbf{+49.1\%}$ & $\mathbf{+50.0\%}$ & $\mathbf{+74.2\%}$ \\
\bottomrule
\end{tabular}
\end{subtable}

\vspace{6pt}
\begin{subtable}{\textwidth}
\centering
\caption{vs.\ LR}
\begin{tabular}{lccccc}
\toprule
\textbf{Model} & Acc & AUCROC & DP & EOD & EOP \\
\midrule
\ftfm{}-0.7 & $+0.1\%$ & $+2.9\%$ & $\mathbf{+28.0\%}$ & $\mathbf{+26.9\%}$ & $\mathbf{+30.9\%}$ \\
\ftfm{}-1.0 & $-0.3\%$ & $+2.5\%$ & $\mathbf{+29.9\%}$ & $\mathbf{+32.1\%}$ & $\mathbf{+31.9\%}$ \\
\ftfm{}-10 & $-2.3\%$ & $+1.4\%$ & $\mathbf{+41.1\%}$ & $\mathbf{+43.3\%}$ & $\mathbf{+45.7\%}$ \\
\ftfm{}-25 & $-5.5\%$ & $-0.3\%$ & $\mathbf{+44.9\%}$ & $\mathbf{+46.3\%}$ & $\mathbf{+73.4\%}$ \\
\bottomrule
\end{tabular}
\end{subtable}
\end{table}
}

\subsection{Performance dynamics during training}\label{app:training-dynamics}

Figure~\ref{fig:dynamics-during-training} \rev{in the main text} provides a finer-grained view of how fairness emerges over the course of pretraining. We track accuracy and fairness (DP and EOD) for \ftfm with $\lambda \in \{0.7, 25\}$ and for nanoTabPFN, evaluating checkpoints throughout training and averaging the resulting curves over 10 real-world fairness tasks from our benchmark. The shaded regions indicate one standard deviation across these tasks. Since lower DP and EOD correspond to fairer behavior, the middle and right panels show that both \ftfm{} variants achieve a fairness advantage over nanoTabPFN very early in training and retain this advantage throughout optimization. This gap is not confined to a narrow set of checkpoints: it persists across most of the training trajectory and becomes especially pronounced in the later stages. \rev{We note that the nanoTabPFN reference is trained under the same data prior, backbone, optimizer, and training budget as \ftfm{}; the two differ only in the fairness-specific components (the sensitive-feature designation in task sampling, the added encoder and adversarial head, and the fairness loss), so this comparison isolates the effect of our intervention.}

At the same time, the figure makes the fairness--accuracy trade-off induced by $\lambda$ visually explicit. Specifically, nanoTabPFN and \ftfm with $\lambda=0.7$ converge to very similar final accuracies, but the latter does so while maintaining consistently lower DP and EOD gaps. By contrast, \ftfm with $\lambda=25$ occupies a distinctly different operating regime: its accuracy remains below that of the other two models, but it attains by far the lowest unfairness on both metrics. The trajectories are also mildly non-monotonic, which is expected because fairness and accuracy are measured on heterogeneous downstream tasks rather than on the pretraining objective itself. Taken together, these dynamics show that our objective changes the optimization path, steering the encoder toward representations that are progressively less informative about the sensitive attribute, with $\lambda$ providing a direct handle on the final fairness--accuracy operating point.

\subsection{Fair representation learning: protocol details and additional results}\label{app:fair-rep-details}

The fairness condition in Definition~\ref{def:fair-representation} is equivalent to $\mathbb{P}(Z \mid Y, S) = \mathbb{P}(Z \mid Y)$ or, in information-theoretic terms, $I(Z; S \mid Y) = 0$. A downstream model trained on $Z$ is therefore expected to exhibit improved fairness relative to a model trained directly on $X$.

Existing approaches for learning $g$ are typically task-specific, requiring a separate model to be fitted for each downstream task~\citep{zemel2013learning,madras2018learning}. We compare \ftfm{} against two open-source, sklearn-compatible pre-processing methods from the \texttt{fairlearn} library~\citep{bird2020fairlearn}. The first is Correlation Remover (CR), which reduces linear dependence on the sensitive attribute by applying a linear transformation to the non-sensitive features~\citep{bird2020fairlearn}. The second is Learning Fair Representations (LFR), which maps inputs to latent prototypes while encouraging similar assignment behaviour across demographic groups~\citep{zemel2013learning}.

We consider ACSIncome for Alabama (AL) with three sensitive attributes---gender, race, and age---yielding three downstream fairness tasks. Each task is randomly split into 80\% training data and 20\% test data. For CR and LFR, the training split is used to fit the representation map $g$, which is then applied to both the training and test features. For \ftfm, we use the training split as context with the test as query, and compute embeddings for both training and test examples from the transformer's output. Importantly, these representations are obtained in a single forward pass, without any task-specific gradient updates.

We train three downstream classifiers---LR, RF, and XGB---on four types of representations: the raw input features and the fair representations produced by CR, LFR, and \ftfm{}. For CR and LFR, we sweep the parameter controlling the fairness--accuracy trade-off over $[0.01, 0.02, \ldots, 0.1, 0.2, \ldots, 1.0]$, with denser coverage in the higher-fairness regime. For \ftfm{}, we evaluate the checkpoints associated with the $\lambda$ values used in Figure~\ref{fig:main-results}. We run this experiment across three random seeds and average the results.

\begin{figure}[h]
    \centering
    \begin{subfigure}{.95\textwidth}
        \centering
        \includegraphics[width=\linewidth]{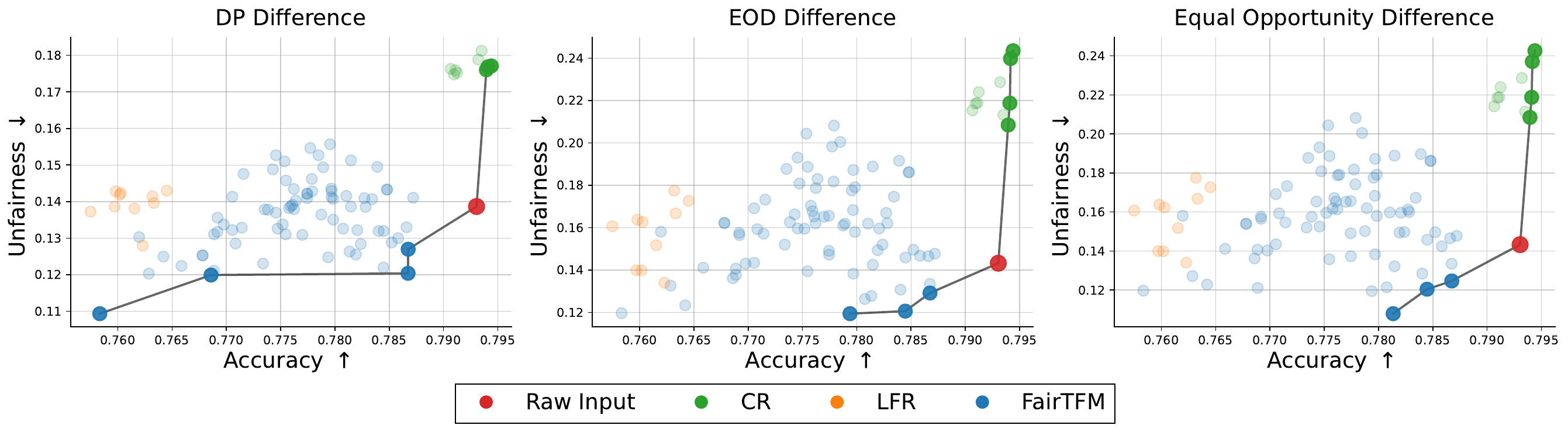}
        \caption{Race}
        \label{fig:fair-rep-XGB-RACE}
    \end{subfigure}

    \begin{subfigure}{.95\textwidth}
        \centering
        \includegraphics[width=\linewidth]{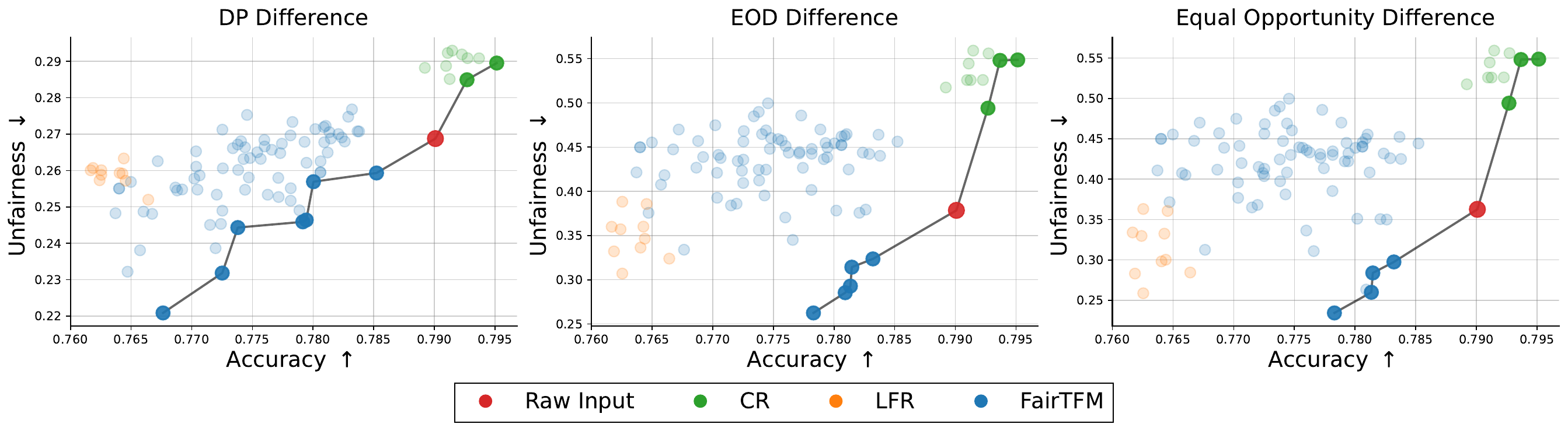}
        \caption{Age}
        \label{fig:fair-rep-XGB-AGE}
    \end{subfigure}

   \caption{Pareto front of XGBoost models trained with different data representations, for race and age as sensitive attributes \rev{(gender is shown in Figure~\ref{fig:fair-rep-XGB-gender} of the main text)}.}
    \label{fig:fair-rep-XGB}
\end{figure}

\begin{figure}[t]
    \centering
    \begin{subfigure}{\textwidth}
        \centering
        \includegraphics[width=\linewidth]{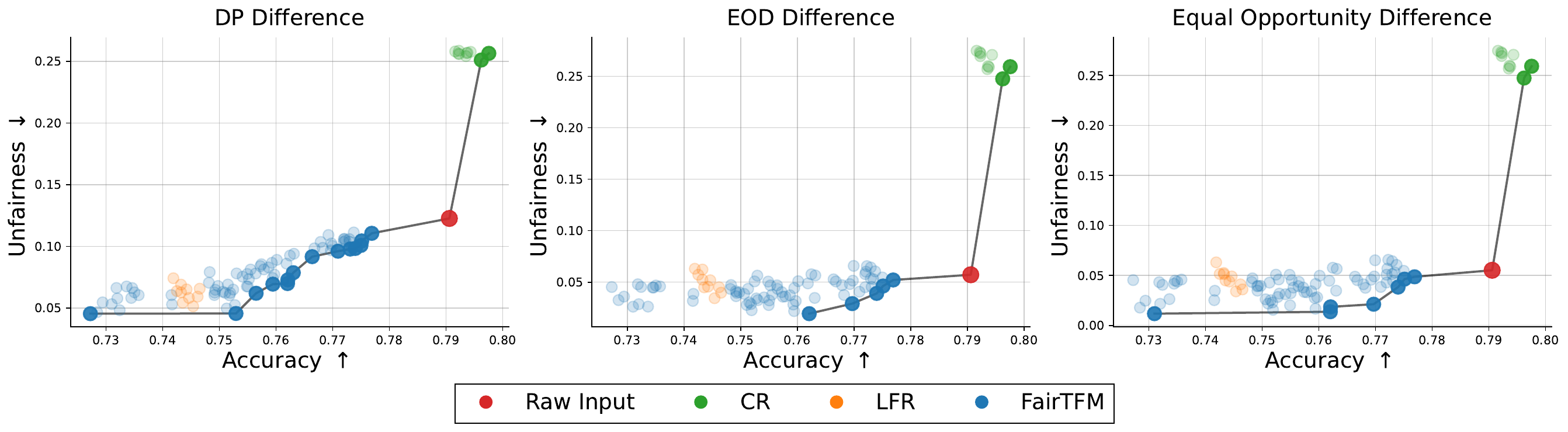}
        \caption{Gender}
        \label{fig:fair-rep-RF-Gender}
    \end{subfigure}
    \hfill
    \begin{subfigure}{\textwidth}
        \centering
        \includegraphics[width=\linewidth]{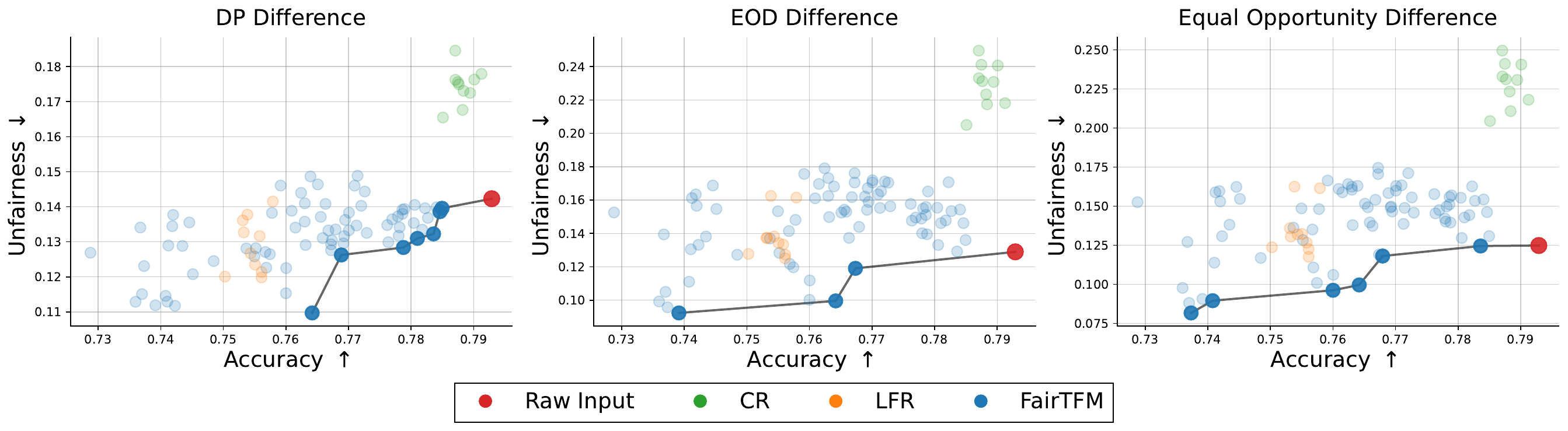}
        \caption{Race}
        \label{fig:fair-rep-RF-RACE}
    \end{subfigure}

    \begin{subfigure}{\textwidth}
        \centering
        \includegraphics[width=\linewidth]{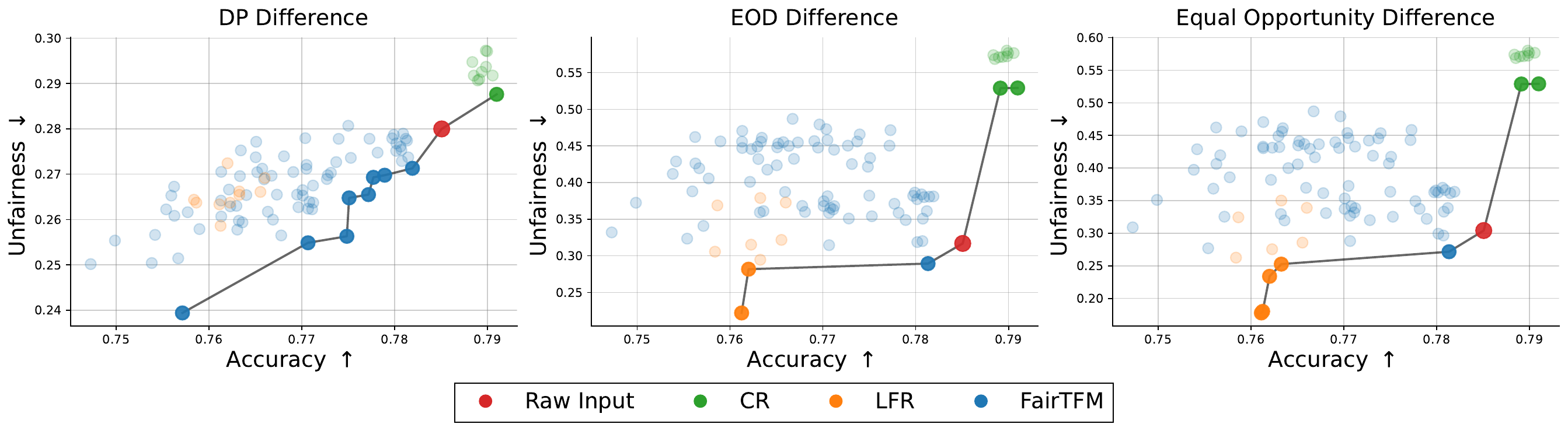}
        \caption{Age}
        \label{fig:fair-rep-RF-AGE}
    \end{subfigure}
    
   \caption{Pareto front of Random Forest models trained with different data representations.}
    \label{fig:fair-rep-RF}
\end{figure}

\begin{figure}[t]
    \centering
    \begin{subfigure}{\textwidth}
        \centering
        \includegraphics[width=\linewidth]{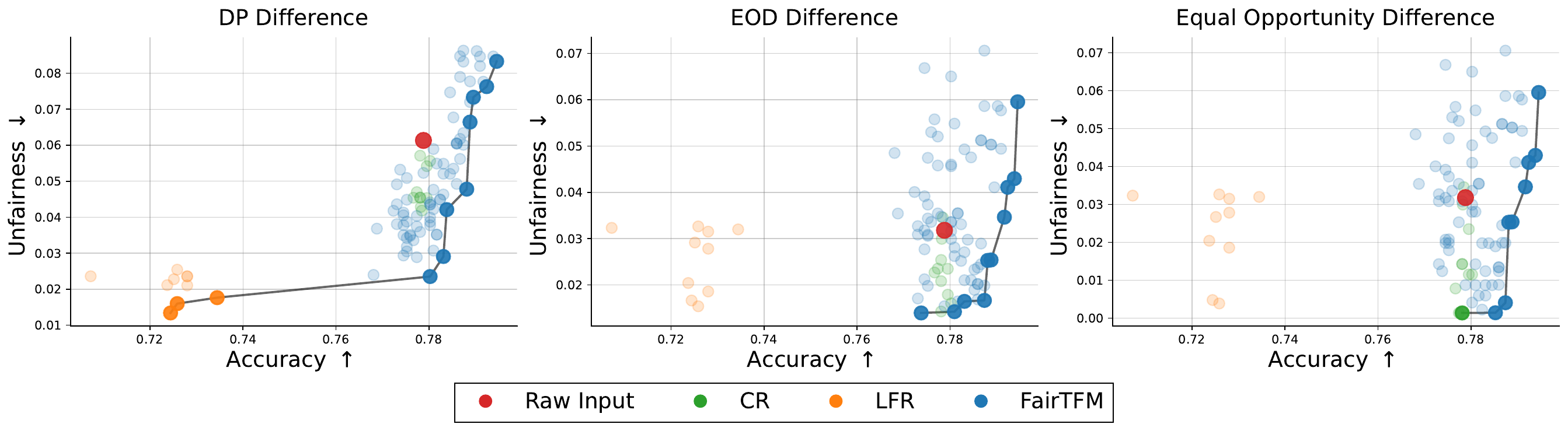}
        \caption{Gender}
        \label{fig:fair-rep-LR-Gender}
    \end{subfigure}
    \hfill
    \begin{subfigure}{\textwidth}
        \centering
        \includegraphics[width=\linewidth]{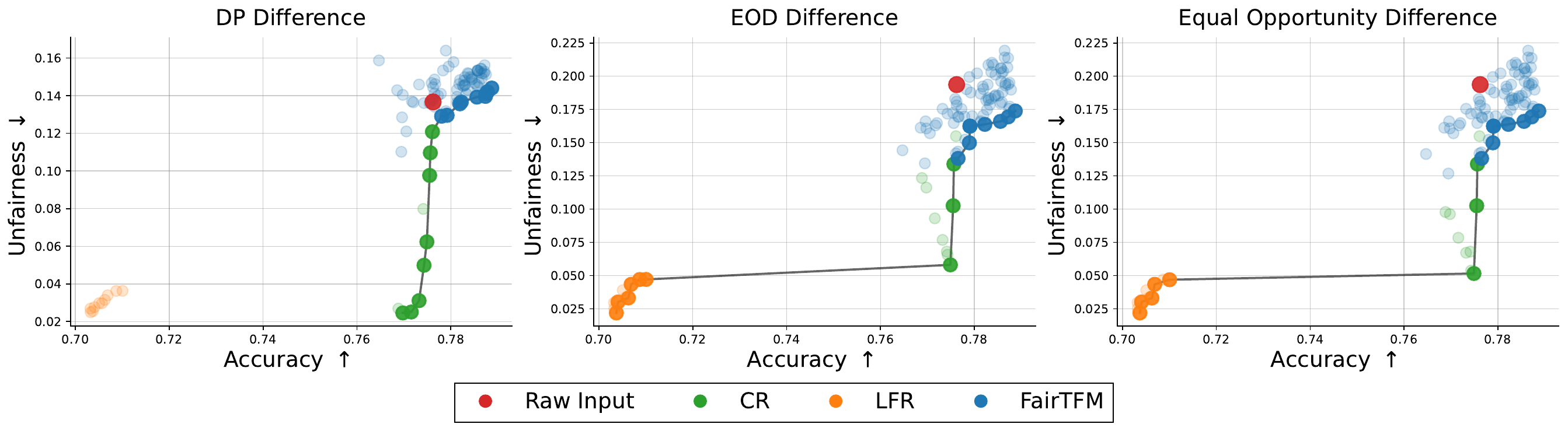}
        \caption{Race}
        \label{fig:fair-rep-LR-RACE}
    \end{subfigure}

    \begin{subfigure}{\textwidth}
        \centering
        \includegraphics[width=\linewidth]{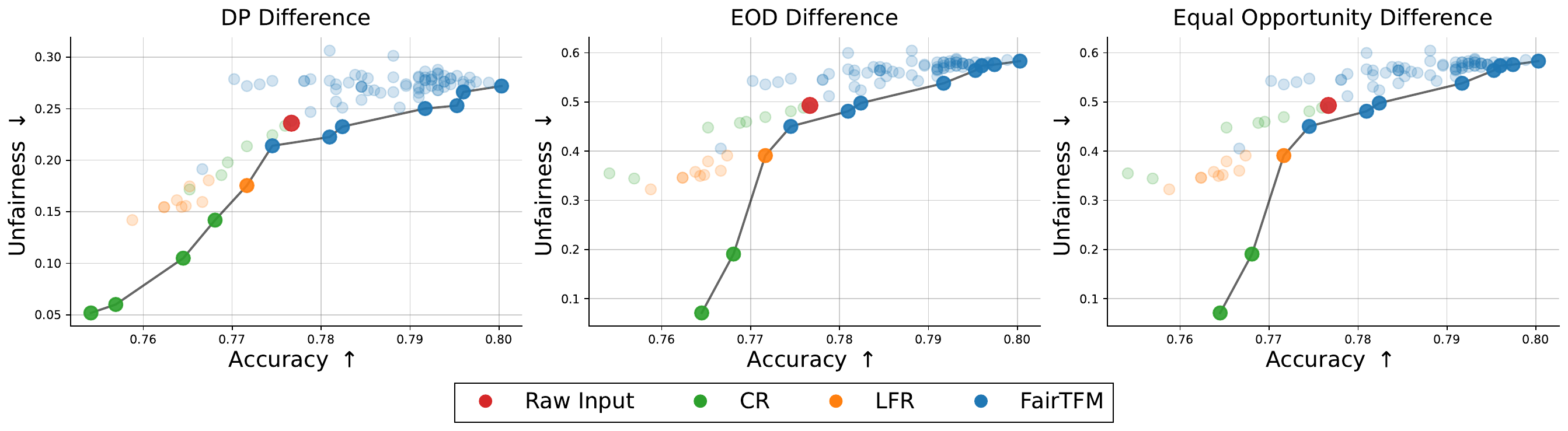}
        \caption{Age}
        \label{fig:fair-rep-LR-AGE}
    \end{subfigure}
    
   \caption{Pareto front of Logistic Regression models trained with different data representations.}
    \label{fig:fair-rep-LR}
\end{figure}

\rev{Figure~\ref{fig:fair-rep-XGB} shows the XGBoost results for the remaining sensitive attributes (race and age; gender is in Figure~\ref{fig:fair-rep-XGB-gender} of the main text), and Figures~\ref{fig:fair-rep-RF} and~\ref{fig:fair-rep-LR} show the Pareto fronts with Random Forest and Logistic Regression as downstream models. Across all three downstream models, \ftfm{} embeddings provide competitive fairness--accuracy trade-offs without any task-specific fitting of the representation map, while CR is most effective for the linear LR model, as it only removes linear correlation with the sensitive attribute.}

\subsection{Additional Exponentiated Gradient results}

\rev{Figure~\ref{fig:exgrad-baseline-race-age} complements Figure~\ref{fig:exgrad-baseline} in the main text by showing the fairness--accuracy trade-off against the EG-constrained baselines when race and age are used as sensitive attributes (same setup, ACSIncome in Alabama). For race, where all methods incur larger fairness gaps, \ftfm{} spans a wide and competitive portion of the frontier, particularly at the lower-unfairness end. For age, EG-based classical models achieve stronger accuracy--fairness trade-offs for EOD and EOP, whereas \ftfm remains competitive on DP.}

\rev{
\begin{figure}[h]
    \centering
    \begin{subfigure}{\textwidth}
        \centering
        \includegraphics[width=.95\linewidth]{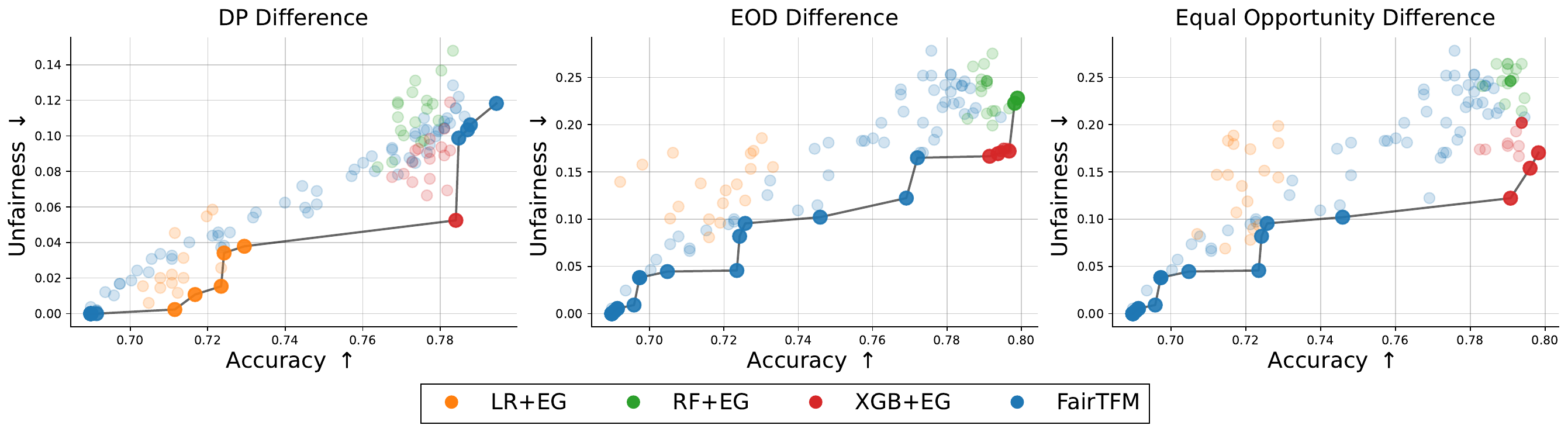}
        \caption{Race}
        \label{fig:exgrad-RACE-1}
    \end{subfigure}

    \begin{subfigure}{\textwidth}
        \centering
        \includegraphics[width=.95\linewidth]{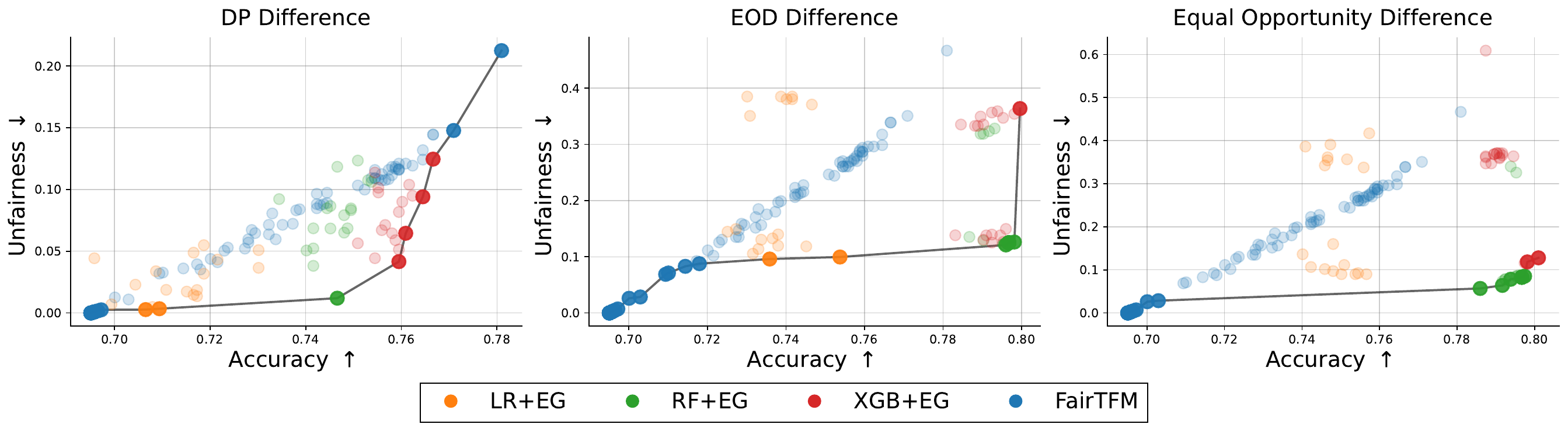}
        \caption{Age}
        \label{fig:exgrad-AGE-1}
    \end{subfigure}

   \caption{Fairness-accuracy trade-off comparison with task-specific classical models trained with fairness constraints using Exponentiated Gradient (EG), on ACSIncome (Alabama) with race and age as sensitive attributes (gender is shown in Figure~\ref{fig:exgrad-baseline} of the main text).}
    \label{fig:exgrad-baseline-race-age}
\end{figure}
}

\begin{figure}[t]
    \centering
    \begin{subfigure}{\textwidth}
        \centering
        \includegraphics[width=\linewidth]{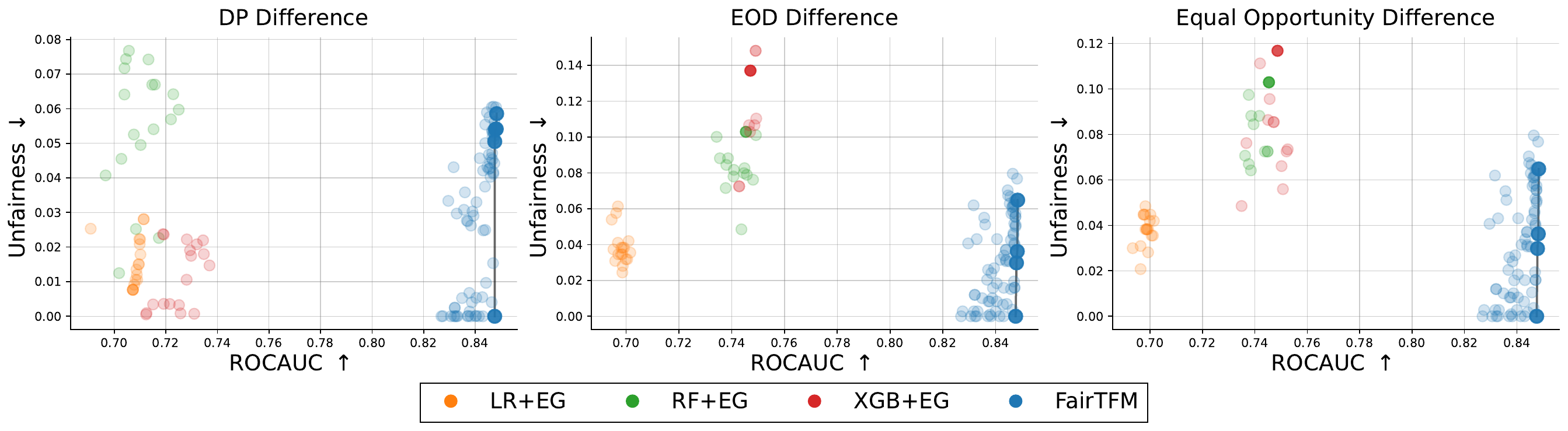}
        \caption{Gender}
        \label{fig:exgrad-Gender}
    \end{subfigure}
    \hfill
    \begin{subfigure}{\textwidth}
        \centering
        \includegraphics[width=\linewidth]{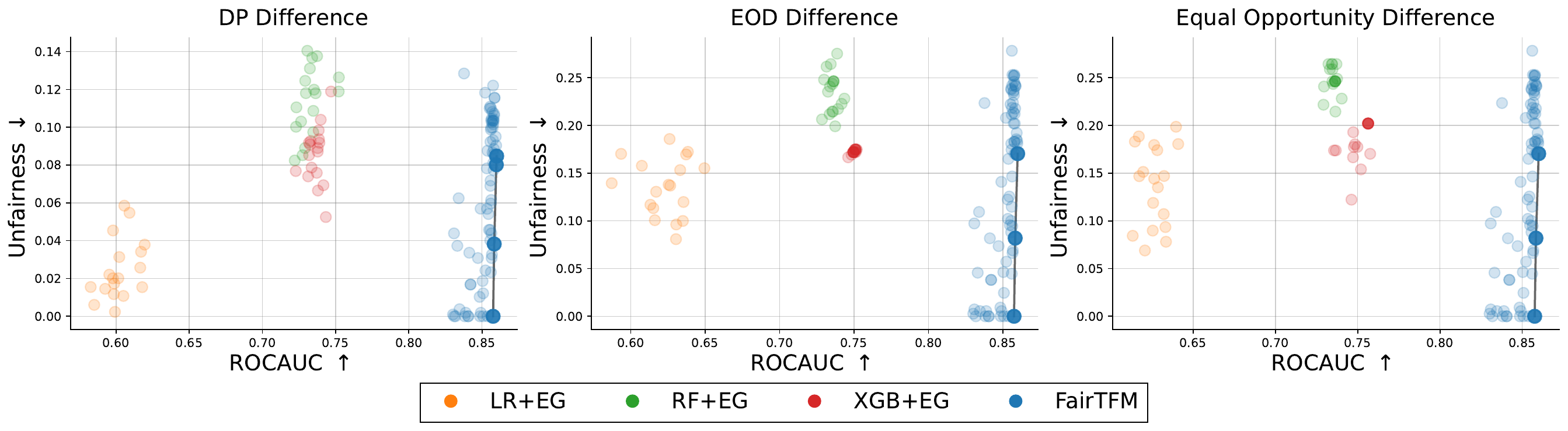}
        \caption{Race}
        \label{fig:exgrad-RACE}
    \end{subfigure}

    \begin{subfigure}{\textwidth}
        \centering
        \includegraphics[width=\linewidth]{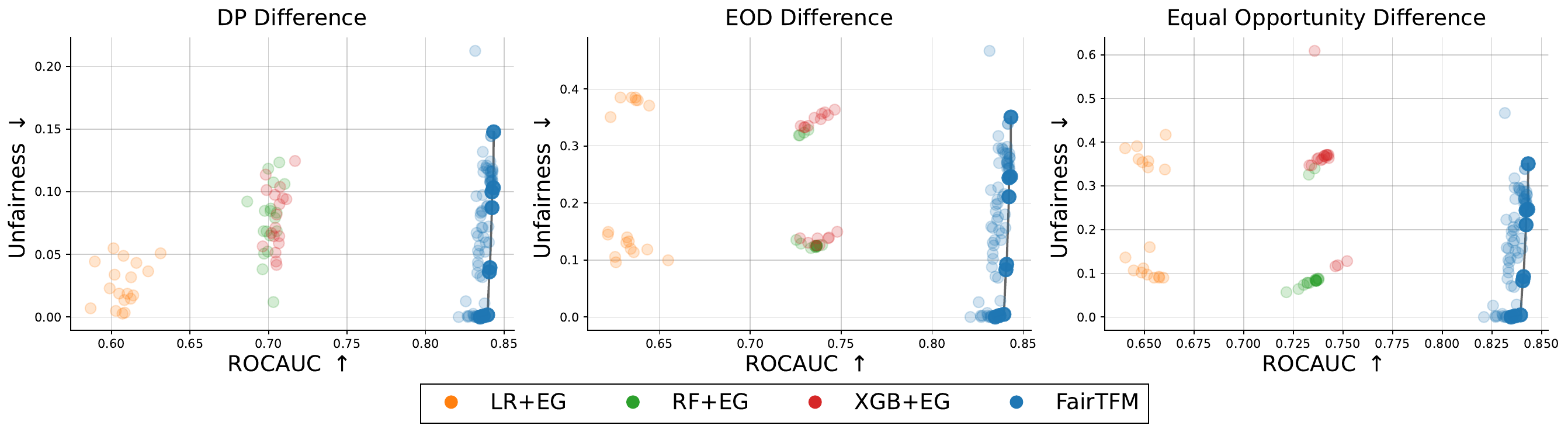}
        \caption{Age}
        \label{fig:exgrad-AGE}
    \end{subfigure}
    
   \caption{Pareto front of classical models trained with fairness constraints using Exponentiated Gradient (EG). \ftfm{} show best overall pareto front when performance are measured with AUCROC.}
    \label{fig:exgrad-baseline-auc}
\end{figure}

\rev{Figure~\ref{fig:exgrad-baseline-auc} additionally measures predictive performance with AUCROC instead of accuracy. Because the EG reduction produces randomized ensembles that trade calibrated scores for constraint satisfaction, its AUCROC drops sharply, and \ftfm{} provides the best overall Pareto front across all three sensitive attributes under this view.}

\clearpage
\rev{
\subsection{Evaluation beyond ACS PUMS tasks}\label{app:other-datasets}

The main evaluation covers 120 tasks derived from ACS PUMS~\citep{ding2021retiring}. To test whether the observed fairness--utility trends extend beyond this data source, we additionally evaluate on a collection of widely used fairness benchmarks spanning different domains, prediction targets, and sensitive attributes~\citep{le2022survey}. Specifically, we consider the following datasets:
\begin{itemize}[leftmargin=*, nosep]
    \item The Adult (Census Income) dataset contains demographic and socioeconomic records from the U.S. Census~\citep{adult_1996}. The task is to predict whether an individual's annual income exceeds \$50,000, using gender, race, and age as sensitive attributes.
    \item The COMPAS dataset contains criminal justice screening records for defendants~\citep{angwin2022machine}. The task is to predict whether an individual will be rearrested within two years of their initial arrest, with race as the sensitive attribute.
    \item The German Credit dataset contains records of bank account holders and is commonly used for credit risk assessment. The task is to classify applicants as low-risk or high-risk, using gender and age ($\leq$25 years) as sensitive attributes.
    \item The Diabetes dataset contains clinical records from 130 U.S. hospitals collected between 1999 and 2008~\citep{strack2014impact}. The task is to predict whether a patient will be readmitted within 30 days of discharge, using gender and race as sensitive attributes.
    \item The Law School dataset consists of admissions records collected by the Law School Admission Council (LSAC) from 163 U.S. law schools in 1991~\citep{wightman1998lsac}. The task is to predict whether a candidate will pass the bar exam, using race and gender as sensitive attributes.
    \item The CelebA dataset provides facial-attribute annotations for celebrity images. We consider two binary prediction tasks, \textit{Blond Hair} and \textit{Smiling}, using gender as the sensitive attribute in both cases.
\end{itemize}

Together, these benchmarks define 12 additional fairness tasks. We use the same evaluation protocol as in Section~\ref{sec:results} and average results over three random seeds. Figure~\ref{fig:result-other-dataset} in the main text shows the accuracy Pareto fronts, Figure~\ref{fig:result-other-dataset-auc} below shows the AUCROC view, and Table~\ref{tab:other-datasets} reports the averaged metrics, playing the same role as Table~\ref{tab:baseline_comparison} for the 120 ACS tasks. As there, the table reports the last pretraining checkpoint of each \ftfm variant, so it is a conservative summary: Pareto-dominant checkpoints in the figures could yield better trade-offs, at the cost of model selection on a real-world validation set. The trend from the main evaluation transfers fully to these tasks: \ftfm{}-0.7 reduces DP by 49\% and EOD by 47\% relative to TabPFNv2.5 while matching XGB and LR in AUCROC (0.803), and \ftfm{}-25 nearly eliminates group disparities (DP of 0.011, an 89\% reduction). Notably, KNN---the fairest unconstrained baseline---achieves its lower disparities only through much weaker predictive performance (AUCROC 0.664), whereas \ftfm offers strictly better fairness at substantially higher AUCROC. Since \ftfm{} is pretrained on purely synthetic data, all real datasets are unseen during pretraining; these results additionally rule out any ACS-specific effect in the evaluation and show that the fairness behavior learned from synthetic pretraining generalizes across data provenances.

\begin{table}[h]
\centering
\caption{Average fairness and accuracy metrics across the 12 fairness tasks beyond ACS PUMS, reported as mean $\pm$ standard deviation over three random seeds. The same pattern as on the 120 ACS tasks holds: unconstrained TFMs are the most accurate but least fair, while \ftfm variants sharply reduce disparities at a moderate predictive cost.}
\label{tab:other-datasets}
\renewcommand{\arraystretch}{1.02}
\setlength{\tabcolsep}{4.5pt}
\begin{tabular}{lccccc}
\toprule
\textbf{Model} & \textbf{Accuracy} & \textbf{AUCROC} & \textbf{DP Diff} & \textbf{EOD Diff} & \textbf{EOP Diff} \\
\midrule
TabICLv2 & $\mathbf{0.796 \pm 0.09}$ & $\mathbf{0.829 \pm 0.09}$ & $0.101 \pm 0.09$ & $0.097 \pm 0.06$ & $0.087 \pm 0.06$ \\
TabPFNv2.5 & $0.793 \pm 0.10$ & $0.828 \pm 0.09$ & $0.097 \pm 0.09$ & $0.106 \pm 0.07$ & $0.092 \pm 0.08$ \\
\midrule
KNN & $0.732 \pm 0.10$ & $0.664 \pm 0.14$ & $0.065 \pm 0.06$ & $0.078 \pm 0.05$ & $0.062 \pm 0.06$ \\
XGB & $0.780 \pm 0.10$ & $0.803 \pm 0.10$ & $0.100 \pm 0.09$ & $0.099 \pm 0.07$ & $0.085 \pm 0.07$ \\
LR & $0.784 \pm 0.09$ & $0.803 \pm 0.10$ & $0.099 \pm 0.09$ & $0.115 \pm 0.08$ & $0.087 \pm 0.07$ \\
RF & $0.786 \pm 0.09$ & $0.806 \pm 0.09$ & $0.102 \pm 0.09$ & $0.112 \pm 0.06$ & $0.084 \pm 0.05$ \\
\midrule
FairPFN & $0.737 \pm 0.10$ & $0.705 \pm 0.10$ & $0.045 \pm 0.08$ & $0.062 \pm 0.09$ & $0.043 \pm 0.08$ \\
\midrule
\ftfm{}-0.1 & $0.753 \pm 0.11$ & $0.782 \pm 0.13$ & $0.052 \pm 0.09$ & $0.060 \pm 0.10$ & $0.060 \pm 0.10$ \\
\ftfm{}-0.7 & $0.752 \pm 0.10$ & $0.803 \pm 0.09$ & $0.049 \pm 0.08$ & $0.056 \pm 0.09$ & $0.056 \pm 0.09$ \\
\ftfm{}-10 & $0.734 \pm 0.10$ & $0.780 \pm 0.10$ & $0.021 \pm 0.07$ & $0.025 \pm 0.07$ & $0.025 \pm 0.07$ \\
\ftfm{}-25 & $0.708 \pm 0.14$ & $0.737 \pm 0.12$ & $\mathbf{0.011 \pm 0.10}$ & $\mathbf{0.015 \pm 0.10}$ & $\mathbf{0.015 \pm 0.10}$ \\
\bottomrule
\end{tabular}
\end{table}

\begin{figure}[h]
    \centering
    \includegraphics[width=.95\linewidth]{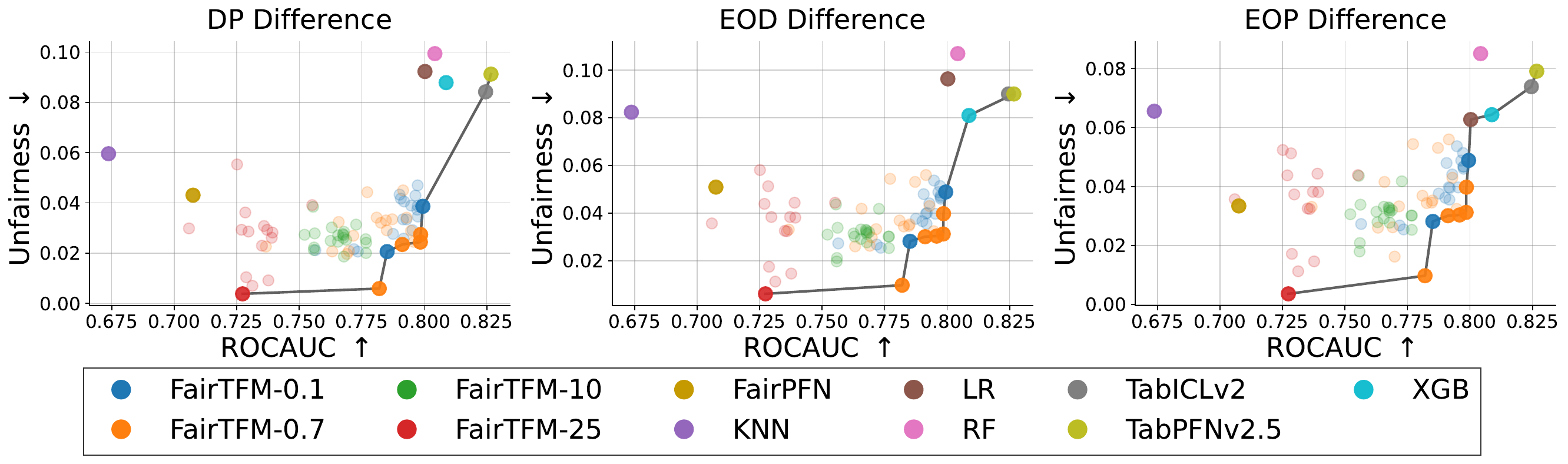}
   \caption{Pareto front between AUCROC and fairness for various models on the 12 fairness tasks beyond ACS PUMS. $\uparrow$ indicates higher is better and $\downarrow$ indicates smaller is better (unfairness).}
    \label{fig:result-other-dataset-auc}
\end{figure}

As for the 120 ACS tasks, Table~\ref{tab:percent-improvement-other} reports the relative change of each \ftfm variant against TabPFNv2.5, XGB, and LR, derived from Table~\ref{tab:other-datasets}. The picture is even stronger than on the ACS benchmark: \ftfm{}-0.7 improves DP, EOD, and EOP by 39--50\% against TabPFNv2.5 for a 5.2\% accuracy and 3.0\% AUCROC cost, and against XGB and LR it improves fairness by 34--51\% with \emph{no} AUCROC cost at all (0.803 for all three models). At the fair end, \ftfm{}-25 improves fairness by 82--89\% across all three baselines.

\begin{table}[h]
\centering
\caption{Fairness improvement (positive numbers mean fairer) and accuracy and AUCROC change of each \ftfm variant relative to (a) TabPFNv2.5, (b) XGB, and (c) LR, averaged over the 12 fairness tasks beyond ACS PUMS and derived from Table~\ref{tab:other-datasets}.}
\label{tab:percent-improvement-other}
\renewcommand{\arraystretch}{1.1}
\setlength{\tabcolsep}{6pt}
\begin{subtable}{\textwidth}
\centering
\caption{vs.\ TabPFNv2.5}
\begin{tabular}{lccccc}
\toprule
\textbf{Model} & Acc & AUCROC & DP & EOD & EOP \\
\midrule
\ftfm{}-0.1 & $-5.0\%$ & $-5.6\%$ & $\mathbf{+46.4\%}$ & $\mathbf{+43.4\%}$ & $\mathbf{+34.8\%}$ \\
\ftfm{}-0.7 & $-5.2\%$ & $-3.0\%$ & $\mathbf{+49.5\%}$ & $\mathbf{+47.2\%}$ & $\mathbf{+39.1\%}$ \\
\ftfm{}-10 & $-7.4\%$ & $-5.8\%$ & $\mathbf{+78.4\%}$ & $\mathbf{+76.4\%}$ & $\mathbf{+72.8\%}$ \\
\ftfm{}-25 & $-10.7\%$ & $-11.0\%$ & $\mathbf{+88.7\%}$ & $\mathbf{+85.8\%}$ & $\mathbf{+83.7\%}$ \\
\bottomrule
\end{tabular}
\end{subtable}

\vspace{6pt}
\begin{subtable}{\textwidth}
\centering
\caption{vs.\ XGB}
\begin{tabular}{lccccc}
\toprule
\textbf{Model} & Acc & AUCROC & DP & EOD & EOP \\
\midrule
\ftfm{}-0.1 & $-3.5\%$ & $-2.6\%$ & $\mathbf{+48.0\%}$ & $\mathbf{+39.4\%}$ & $\mathbf{+29.4\%}$ \\
\ftfm{}-0.7 & $-3.6\%$ & $\pm0.0\%$ & $\mathbf{+51.0\%}$ & $\mathbf{+43.4\%}$ & $\mathbf{+34.1\%}$ \\
\ftfm{}-10 & $-5.9\%$ & $-2.9\%$ & $\mathbf{+79.0\%}$ & $\mathbf{+74.7\%}$ & $\mathbf{+70.6\%}$ \\
\ftfm{}-25 & $-9.2\%$ & $-8.2\%$ & $\mathbf{+89.0\%}$ & $\mathbf{+84.8\%}$ & $\mathbf{+82.4\%}$ \\
\bottomrule
\end{tabular}
\end{subtable}

\vspace{6pt}
\begin{subtable}{\textwidth}
\centering
\caption{vs.\ LR}
\begin{tabular}{lccccc}
\toprule
\textbf{Model} & Acc & AUCROC & DP & EOD & EOP \\
\midrule
\ftfm{}-0.1 & $-4.0\%$ & $-2.6\%$ & $\mathbf{+47.5\%}$ & $\mathbf{+47.8\%}$ & $\mathbf{+31.0\%}$ \\
\ftfm{}-0.7 & $-4.1\%$ & $\pm0.0\%$ & $\mathbf{+50.5\%}$ & $\mathbf{+51.3\%}$ & $\mathbf{+35.6\%}$ \\
\ftfm{}-10 & $-6.4\%$ & $-2.9\%$ & $\mathbf{+78.8\%}$ & $\mathbf{+78.3\%}$ & $\mathbf{+71.3\%}$ \\
\ftfm{}-25 & $-9.7\%$ & $-8.2\%$ & $\mathbf{+88.9\%}$ & $\mathbf{+87.0\%}$ & $\mathbf{+82.8\%}$ \\
\bottomrule
\end{tabular}
\end{subtable}
\end{table}

\clearpage
\subsection{Fairness-aware baselines on the 12 non-ACS tasks}\label{app:eg-other-datasets}

We next compare \ftfm{} against task-specific fairness-aware baselines on the same 12 non-ACS tasks. We wrap XGBoost with the Exponentiated Gradient (EG) reduction~\citep{agarwal2018reductions}; we focus on XGBoost because it is among the strongest non-TFM baselines on these tasks (Table~\ref{tab:other-datasets}) and was the strongest EG-wrapped model in the ACS comparison (Figure~\ref{fig:exgrad-baseline}). For each of the three fairness constraints (DP, EOD, EOP), we sweep the fairness-violation tolerance $\varepsilon$ over $\{0.01, 0.02, 0.03, 0.05, 0.07, 0.10, 0.20, 0.30, 0.50, 0.70, 0.90\}$, yielding 33 configurations, each retrained per task and averaged over three seeds. Table~\ref{tab:expgrad_comparison} reports all configurations, and Table~\ref{tab:eg-summary} summarizes the comparison using, for each constraint type, the best (lowest) value of its targeted metric across $\varepsilon$.

Three observations follow. First, each \ftfm variant improves on the fairness metric that each EG variant explicitly optimizes: relative to the best EG value of each targeted metric (DP 0.062, EOD 0.083, EOP 0.070), \ftfm{}-0.7 improves by 21\% (DP), 33\% (EOD), and 20\% (EOP), \ftfm{}-10 by 66\%, 70\%, and 64\%, and \ftfm{}-25 by 82\%, 82\%, and 79\%. Second, this comes at substantially higher predictive performance: the EG randomized ensembles reduce XGBoost's AUCROC from 0.803 to roughly 0.69--0.70 regardless of $\varepsilon$, while \ftfm{}-0.7 keeps AUCROC at 0.803 with only a modest accuracy cost. Third, the comparison highlights a structural advantage: obtaining the EG numbers required retraining $33 \times 12 \times 3$ model configurations, whereas a single pretrained \ftfm checkpoint serves every task and constraint in one forward pass. This per-task, per-constraint retraining cost is itself part of the case for fairness-aware pretraining.

\begin{table}[h]
\centering 
\caption{Summary comparison on the 12 non-ACS tasks. For each EG constraint type, we report the configuration achieving the best (lowest) value of its targeted metric across the $\varepsilon$ sweep (targeted metric in bold); unconstrained XGBoost is included as a reference. Each \ftfm variant is fairer on every targeted metric at substantially higher AUCROC.}
\label{tab:eg-summary}
\renewcommand{\arraystretch}{1.05}
\setlength{\tabcolsep}{5pt}
\begin{tabular}{lccccc}
\toprule
\textbf{Model} & \textbf{Accuracy} & \textbf{AUCROC} & \textbf{DP Diff} & \textbf{EOD Diff} & \textbf{EOP Diff} \\
\midrule
XGB (unconstrained) & $0.780$ & $0.803$ & $0.100$ & $0.099$ & $0.085$ \\
XGB-EG-DP (best $\varepsilon$) & $0.777$ & $0.695$ & $\mathbf{0.062}$ & $0.075$ & $0.060$ \\
XGB-EG-EOD (best $\varepsilon$) & $0.778$ & $0.702$ & $0.084$ & $\mathbf{0.083}$ & $0.068$ \\
XGB-EG-EOP (best $\varepsilon$) & $0.777$ & $0.700$ & $0.085$ & $0.084$ & $\mathbf{0.070}$ \\
\midrule
\ftfm{}-0.7 & $0.752$ & $0.803$ & $0.049$ & $0.056$ & $0.056$ \\
\ftfm{}-10 & $0.734$ & $0.780$ & $0.021$ & $0.025$ & $0.025$ \\
\ftfm{}-25 & $0.708$ & $0.737$ & $0.011$ & $0.015$ & $0.015$ \\
\bottomrule
\end{tabular}
\end{table}

\begin{table}[h]
\centering
\caption{Full Exponentiated Gradient results on the 12 non-ACS tasks: average fairness and accuracy metrics for each fairness constraint (DP, EOD, EOP) and tolerance $\varepsilon$, mean $\pm$ standard deviation over three random seeds.}
\label{tab:expgrad_comparison}
\renewcommand{\arraystretch}{1.0}
\setlength{\tabcolsep}{4pt}
\small
\begin{tabular}{clccccc}
\toprule
\textbf{Model} & $\boldsymbol{\varepsilon}$ & \textbf{Accuracy} & \textbf{AUCROC} & \textbf{DP Diff} & \textbf{EOD Diff} & \textbf{EOP Diff} \\
\midrule
\multirow{11}{*}{XGB-EG-DP} & 0.01 & 0.778 $\pm$ 0.10 & 0.695 $\pm$ 0.07 & 0.068 $\pm$ 0.06 & 0.073 $\pm$ 0.03 & 0.061 $\pm$ 0.03 \\
 & 0.02 & 0.778 $\pm$ 0.10 & 0.695 $\pm$ 0.07 & 0.064 $\pm$ 0.05 & 0.065 $\pm$ 0.03 & 0.054 $\pm$ 0.03 \\
 & 0.03 & 0.778 $\pm$ 0.10 & 0.694 $\pm$ 0.07 & 0.064 $\pm$ 0.05 & 0.069 $\pm$ 0.03 & 0.058 $\pm$ 0.03 \\
 & 0.05 & 0.778 $\pm$ 0.10 & 0.692 $\pm$ 0.06 & 0.067 $\pm$ 0.05 & 0.070 $\pm$ 0.03 & 0.059 $\pm$ 0.03 \\
 & 0.07 & 0.777 $\pm$ 0.10 & 0.694 $\pm$ 0.06 & 0.062 $\pm$ 0.05 & 0.072 $\pm$ 0.03 & 0.062 $\pm$ 0.03 \\
 & 0.10 & 0.777 $\pm$ 0.10 & 0.692 $\pm$ 0.06 & 0.068 $\pm$ 0.05 & 0.071 $\pm$ 0.02 & 0.059 $\pm$ 0.03 \\
 & 0.20 & 0.778 $\pm$ 0.10 & 0.693 $\pm$ 0.06 & 0.067 $\pm$ 0.05 & 0.075 $\pm$ 0.04 & 0.061 $\pm$ 0.04 \\
 & 0.30 & 0.776 $\pm$ 0.10 & 0.693 $\pm$ 0.07 & 0.068 $\pm$ 0.06 & 0.078 $\pm$ 0.03 & 0.065 $\pm$ 0.03 \\
 & 0.50 & 0.777 $\pm$ 0.10 & 0.695 $\pm$ 0.07 & 0.062 $\pm$ 0.05 & 0.075 $\pm$ 0.04 & 0.060 $\pm$ 0.04 \\
 & 0.70 & 0.778 $\pm$ 0.10 & 0.696 $\pm$ 0.07 & 0.072 $\pm$ 0.06 & 0.083 $\pm$ 0.04 & 0.068 $\pm$ 0.04 \\
 & 0.90 & 0.777 $\pm$ 0.10 & 0.695 $\pm$ 0.06 & 0.066 $\pm$ 0.06 & 0.071 $\pm$ 0.04 & 0.059 $\pm$ 0.04 \\
\midrule
\multirow{11}{*}{XGB-EG-EOD} & 0.01 & 0.778 $\pm$ 0.10 & 0.702 $\pm$ 0.07 & 0.084 $\pm$ 0.08 & 0.084 $\pm$ 0.05 & 0.069 $\pm$ 0.05 \\
 & 0.02 & 0.778 $\pm$ 0.10 & 0.702 $\pm$ 0.07 & 0.084 $\pm$ 0.08 & 0.083 $\pm$ 0.05 & 0.068 $\pm$ 0.05 \\
 & 0.03 & 0.778 $\pm$ 0.10 & 0.704 $\pm$ 0.07 & 0.084 $\pm$ 0.08 & 0.086 $\pm$ 0.05 & 0.071 $\pm$ 0.05 \\
 & 0.05 & 0.778 $\pm$ 0.10 & 0.704 $\pm$ 0.07 & 0.086 $\pm$ 0.08 & 0.090 $\pm$ 0.05 & 0.075 $\pm$ 0.05 \\
 & 0.07 & 0.778 $\pm$ 0.10 & 0.703 $\pm$ 0.07 & 0.086 $\pm$ 0.08 & 0.088 $\pm$ 0.05 & 0.074 $\pm$ 0.05 \\
 & 0.10 & 0.778 $\pm$ 0.10 & 0.703 $\pm$ 0.07 & 0.085 $\pm$ 0.08 & 0.084 $\pm$ 0.05 & 0.067 $\pm$ 0.05 \\
 & 0.20 & 0.778 $\pm$ 0.10 & 0.700 $\pm$ 0.07 & 0.086 $\pm$ 0.08 & 0.087 $\pm$ 0.06 & 0.072 $\pm$ 0.06 \\
 & 0.30 & 0.779 $\pm$ 0.10 & 0.700 $\pm$ 0.07 & 0.085 $\pm$ 0.07 & 0.084 $\pm$ 0.06 & 0.069 $\pm$ 0.05 \\
 & 0.50 & 0.777 $\pm$ 0.10 & 0.701 $\pm$ 0.07 & 0.084 $\pm$ 0.08 & 0.089 $\pm$ 0.06 & 0.076 $\pm$ 0.06 \\
 & 0.70 & 0.777 $\pm$ 0.10 & 0.700 $\pm$ 0.07 & 0.086 $\pm$ 0.08 & 0.087 $\pm$ 0.06 & 0.074 $\pm$ 0.06 \\
 & 0.90 & 0.778 $\pm$ 0.10 & 0.701 $\pm$ 0.07 & 0.088 $\pm$ 0.08 & 0.088 $\pm$ 0.06 & 0.074 $\pm$ 0.06 \\
\midrule
\multirow{11}{*}{XGB-EG-EOP} & 0.01 & 0.776 $\pm$ 0.10 & 0.698 $\pm$ 0.07 & 0.087 $\pm$ 0.08 & 0.089 $\pm$ 0.06 & 0.074 $\pm$ 0.06 \\
 & 0.02 & 0.776 $\pm$ 0.10 & 0.698 $\pm$ 0.07 & 0.085 $\pm$ 0.08 & 0.087 $\pm$ 0.06 & 0.072 $\pm$ 0.06 \\
 & 0.03 & 0.776 $\pm$ 0.10 & 0.699 $\pm$ 0.07 & 0.086 $\pm$ 0.08 & 0.088 $\pm$ 0.06 & 0.072 $\pm$ 0.06 \\
 & 0.05 & 0.777 $\pm$ 0.10 & 0.699 $\pm$ 0.07 & 0.084 $\pm$ 0.08 & 0.087 $\pm$ 0.06 & 0.073 $\pm$ 0.06 \\
 & 0.07 & 0.776 $\pm$ 0.10 & 0.699 $\pm$ 0.07 & 0.085 $\pm$ 0.08 & 0.086 $\pm$ 0.06 & 0.073 $\pm$ 0.06 \\
 & 0.10 & 0.777 $\pm$ 0.10 & 0.699 $\pm$ 0.07 & 0.087 $\pm$ 0.08 & 0.087 $\pm$ 0.06 & 0.074 $\pm$ 0.06 \\
 & 0.20 & 0.778 $\pm$ 0.10 & 0.699 $\pm$ 0.07 & 0.086 $\pm$ 0.08 & 0.085 $\pm$ 0.06 & 0.071 $\pm$ 0.06 \\
 & 0.30 & 0.777 $\pm$ 0.10 & 0.700 $\pm$ 0.07 & 0.085 $\pm$ 0.09 & 0.084 $\pm$ 0.06 & 0.070 $\pm$ 0.06 \\
 & 0.50 & 0.777 $\pm$ 0.10 & 0.700 $\pm$ 0.07 & 0.088 $\pm$ 0.08 & 0.088 $\pm$ 0.06 & 0.075 $\pm$ 0.06 \\
 & 0.70 & 0.778 $\pm$ 0.10 & 0.699 $\pm$ 0.07 & 0.091 $\pm$ 0.09 & 0.092 $\pm$ 0.06 & 0.077 $\pm$ 0.06 \\
 & 0.90 & 0.777 $\pm$ 0.10 & 0.701 $\pm$ 0.07 & 0.086 $\pm$ 0.08 & 0.091 $\pm$ 0.06 & 0.080 $\pm$ 0.06 \\
\bottomrule
\end{tabular}
\end{table}

}

\end{document}